\documentclass{article}

\usepackage{PRIMEarxiv}
\usepackage[utf8]{inputenc}
\usepackage[T1]{fontenc}
\usepackage{cite}
\usepackage{booktabs}
\usepackage{tabularx}
\usepackage{amsfonts}
\usepackage{amssymb}
\usepackage{amsmath}
\usepackage{amsthm}
\DeclareMathAlphabet{\mathcal}{OMS}{cmsy}{m}{n}
\SetMathAlphabet{\mathcal}{bold}{OMS}{cmsy}{b}{n}
\usepackage{nicefrac}
\usepackage{microtype}
\usepackage{graphicx}
\usepackage{multirow}
\usepackage{fancyhdr}
\usepackage{url}
\usepackage[hidelinks]{hyperref}
\hypersetup{
  pdftitle={Where Does the Watermark Hide? Push–Pull Disentanglement for Invisible Watermark Removal},
  pdfauthor={Anonymous Authors},
  pdfkeywords={image watermarking, watermark removal, paired supervision, push--pull learning, latent-space attack}
}

\renewenvironment{figure*}[1][htbp]{\begin{figure}[htbp]}{\end{figure}}
\renewenvironment{table*}[1][htbp]{\begin{table}[htbp]}{\end{table}}

\providecommand{\citep}[1]{\cite{#1}}
\providecommand{\citet}[1]{\cite{#1}}

\theoremstyle{definition}

\theoremstyle{plain}

\theoremstyle{remark}

\graphicspath{{figures/}}
\title{Where Does the Watermark Hide? Push–Pull Disentanglement for Invisible Watermark Removal}

\author{
Jidong Yang\textsuperscript{1},
Huaike Yu\textsuperscript{1},
Qi Li\textsuperscript{1,*},
Chunpeng Wang\textsuperscript{1},\\
Yuantian Miao\textsuperscript{2},
Suo Gao\textsuperscript{3},
and Xiao Chen\textsuperscript{4}\\[0.5em]
\parbox{0.96\textwidth}{\normalfont\footnotesize\centering
\textsuperscript{1} Jidong Yang, Huaike Yu, Qi Li, and Chunpeng Wang are with the Key Laboratory of Computing Power Network and Information Security, Ministry of Education, Shandong Computer Science Center, and Shandong Provincial Key Laboratory of Industrial Network and Information System Security, Shandong Fundamental Research Center for Computer Science, Qilu University of Technology (Shandong Academy of Sciences), Jinan 250353, China (e-mail: jidong\_yang\_paper@163.com, huaikeyu@gmail.com, qluliqi@163.com, mpeng1122@163.com).\\
\textsuperscript{2} Yuantian Miao is with the Department of Computer Science, City University of Hong Kong (Dongguan), Dongguan, Guangdong 518057, China (e-mail: yuantian.miao@cityu-dg.edu.cn).\\
\textsuperscript{3} Suo Gao is with the School of Information Science and Engineering, Dalian Polytechnic University, Dalian 116034, China (e-mail: gaosuodlpu@163.com).\\
\textsuperscript{4} Xiao Chen is with the School of Computer and Information Sciences, College of Engineering, Science and Environment, The University of Newcastle, Callaghan, NSW 2308, Australia (e-mail: xiao.chen@newcastle.edu.au).\\
\textsuperscript{*} Corresponding author: Qi Li (e-mail: qluliqi@163.com).
}}

\begin{document}

\maketitle

\begin{abstract}
Fixed image distortions do not cover an attacker that learns from paired clean and watermarked images. We study this paired-training threat with single-image inference: deployment uses neither the clean reference nor the watermark key, payload, or decoder. An encoder maps each image to a structural latent $g$ and an auxiliary residual latent $u$. Push supervision reconstructs the watermarked image from $D(g_w,u_w)$. Pull supervision trains the zero-auxiliary output $D(A_g(g_w;k),0)$ toward the paired clean image. At $k=1.10,u=0$, the four-method sweep gives an average BER of $0.3958$, PSNR of $31.07$ dB, and SSIM of $0.9554$. Restoring $u$ from $0$ to $0.15$ moves average BER from $0.3893$ to $0.3357$, while PSNR falls from $30.99$ to $28.23$ dB. The intervention supports decoder dependence on the auxiliary input in the evaluated setting. The accompanying theory is a conditional, post-hoc account of this behavior rather than an experimentally verified information-relocation result.
\end{abstract}


\section{Introduction}
\label{sec:intro}

Invisible image watermarks support provenance and ownership claims only when they remain recoverable after plausible attacks. Current evaluations cover compression, filtering, geometric edits, and generative reconstruction~\citep{an2024waves,zhao2024provably,liu2025ctrlregen}. They do not fully characterize an attacker that learns from aligned clean and watermarked images. Such pairs reveal the image-specific change introduced by a watermarking pipeline and can supervise a removal model that later processes one image at a time.

Paired supervision is already a documented attack resource. The First-Place Solution to the NeurIPS 2024 Invisible Watermark Removal Challenge created 1,000 paired StegaStamp examples and fine-tuned a VAE for single-image removal~\citep{shamshad2025firstplace}. We therefore make no claim that paired training or single-image removal is new. The open question addressed here is narrower: can paired supervision train two decoder inputs with opposing reconstruction targets, then use a zero-auxiliary intervention to obtain a high-fidelity removal path? This formulation exposes an internal control variable that ordinary restoration and regeneration attacks do not provide.

Our model encodes an image as a one-channel structural latent $g$ and a 16-channel auxiliary latent $u$. The push path $D(g_w,u_w)$ reconstructs the watermarked input. The pull path is defined by zeroing the auxiliary input, first conceptually as $D(g_w,0)$ and in the deployed attack as $D(A_g(g_w;k),0)$. Paired clean targets supervise this path, while $A_g$ applies a bandwise frequency perturbation to $g_w$. Training uses pairs; inference uses only the watermarked image. The clean reference, key, payload, and watermark decoder or detector are absent at deployment.

The residual scale supplies a direct intervention on the proposed mechanism. If decoder-relevant watermark evidence depends on $u$, restoring $\alpha u_w$ while holding $A_g(g_w;k)$ fixed should move bit recovery away from random guessing. The experiment follows this pattern. Average BER moves from $0.3893$ at $u=0$ to $0.3357$ at $u=0.15$, while PSNR decreases from $30.99$ to $28.23$ dB. At the selected point $k=1.10,u=0$, the four-method sweep gives BER $0.3958$, PSNR $31.07$ dB, and SSIM $0.9554$. These results motivate a branch-dependence account without asserting that $u$ exclusively contains watermark information.

The paper makes three contributions:
\begin{itemize}
  \item We specify a paired-training, single-image-inference threat model. The training data contain four seen watermark families, while three excluded families test cross-family transfer. Evaluator outputs are used only for offline reporting and operating-point selection.
  \item We introduce a G/U push--pull attack whose full path reconstructs $x_w$ and whose zero-auxiliary path is supervised toward $x_c$. A latent-frequency operator acts on the remaining structural input at inference.
  \item We evaluate the attack on seven watermarking methods. The selected setting reaches average BER $0.3958$ at $31.07$ dB PSNR on the four-method sweep, and the $u$ intervention provides preliminary mechanism evidence. Full scans, per-method results, theory, and implementation details appear in the supplementary material.
\end{itemize}

\section{Related Work}
\label{sec:related}

\subsection{Watermarking and robustness evaluation}
Transform-domain methods embed signals in DCT, DWT, or hybrid components~\citep{barni1998dct,kumari2023dwtsvd}. Neural methods learn the embedding and recovery functions; RivaGAN and StegaStamp are representative examples~\citep{zhang2019rivagan,tancik2020stegastamp}. Generative approaches place structured signals in diffusion sampling or latent decoders, as in Tree-Ring and Stable Signature~\citep{wen2023treering,fernandez2023stable}. These families expose different recovery interfaces. Multi-bit systems return a payload, whereas detector systems return a score or decision. A removal evaluation must preserve this distinction and report image fidelity beside attack success~\citep{an2024waves}.

\subsection{Learned watermark removal}
Fixed distortions test sensitivity to JPEG, blur, noise, or geometry. Regeneration attacks instead damage watermark-bearing components and reconstruct the image with a denoiser or generative prior~\citep{zhao2024provably}. CtrlRegen adds controllable reconstruction to this process~\citep{liu2025ctrlregen}, and SADRE uses saliency-aware reconstruction~\citep{alam2025sadre}. Model-side fine-tuning can also weaken decoder-rooted diffusion watermarks~\citep{hu2024stableunstable}. The ICLR 2025 workshop First-Place Solution is the closest paired-supervised precedent because it trains on 1,000 StegaStamp pairs and removes the watermark from a single test image~\citep{shamshad2025firstplace}. Our difference is the explicit G/U intervention: the same decoder is trained to preserve the watermarked image with $u_w$ and approach the paired clean target when $u$ is zero. The latent-frequency path then perturbs the remaining structural input. This is a mechanism and representation distinction, not a priority claim about paired supervision.

\section{Threat Model}
\label{sec:threat-model}

The attacker receives a training set of paired images and learns a single-image mapping:
\begin{equation}
  \mathcal{D}_{\mathrm{train}}=\{(x_c^{(i)},x_w^{(i)})\}_{i=1}^{N},
  \qquad x_{\mathrm{atk}}=A_\theta(x_w).
  \label{eq:threat}
\end{equation}
The clean image $x_c$ is a training target, not a test-time input. Deployment does not use the key, payload, random seed, or watermark decoder/detector. Offline evaluation does use the relevant decoder or detector to measure removal and select the reported $(k,u)$ point. We describe the deployed mapping as query-free, not the full experimental selection procedure as decoder-free.

The paired training set includes DwtDctSvd, RivaGAN, StegaStamp, and Tree-Ring. These are the \emph{seen families}. DwtDct, SSL Watermarking, and Stable Signature are excluded from paired training and are the \emph{unseen families}. ``Unseen'' refers only to family exclusion. It does not imply new image distributions, payload lengths, keys, resolutions, or implementations.

\begin{table*}[t]
  \caption{Attacker capability by phase. Evaluator access is separated from the deployment inputs.}
  \label{tab:threat-model}
  \centering
  \small
  \begin{tabularx}{\linewidth}{p{0.22\linewidth}p{0.34\linewidth}X}
    \toprule
    Resource & Training and offline selection & Deployment on one image \\
    \midrule
    Image input & Paired $(x_c,x_w)$ examples & One watermarked image $x_w$ \\
    Watermark families & Four represented families & The trained mapping can be applied to any input; results distinguish seen from excluded families \\
    Clean reference & Used as the supervised target & Unavailable \\
    Key, payload, or seed & Not required by the attack training loss & Unavailable \\
    Decoder or detector & Used after training to report metrics and select $(k,u)$ & Not queried or supplied to $A_\theta$ \\
    Objective & Learn the paired clean target while retaining reconstruction fidelity & Produce one attacked image under the fixed selected setting \\
    \bottomrule
  \end{tabularx}
\end{table*}

For a bit decoder, removal is stronger when BER approaches $0.5$. We use
\begin{equation}
  \mathrm{RR}=1-2|\mathrm{BER}-0.5|,
  \label{eq:rr}
\end{equation}
so a deterministic bit inversion at BER one is not counted as evidence destruction. Tree-Ring is evaluated separately: lower TPR at the nominal 1\%-FPR calibration point is better, and Acc Removal is $1-\mathrm{TPR}$. PSNR and SSIM are measured against $x_w$. The attack objective is high removal subject to visible fidelity, rather than decoder failure at any image distortion. Appendix~\ref{app:formal} gives the formal risk definition.

\section{Paired Push--Pull Attack}
\label{sec:method}

\begin{figure*}[t]
  \centering
  \includegraphics[width=\textwidth]{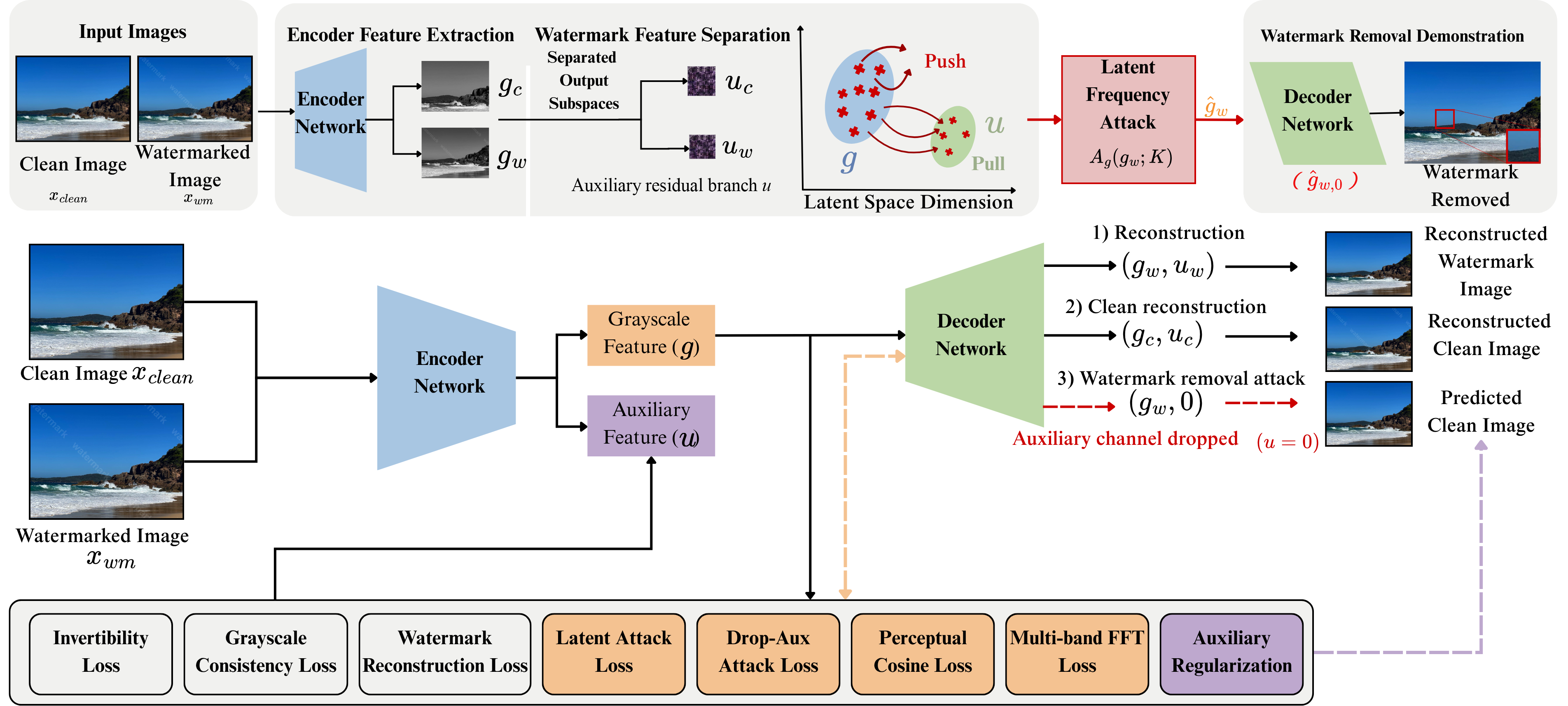}
  \caption{Paired push--pull training and single-image inference. Push reconstructs the watermarked input from $(g_w,u_w)$. Pull supervises the attacked zero-auxiliary output toward the paired clean target. Deployment follows $x_w\rightarrow E\rightarrow(g_w,u_w)\rightarrow(A_g(g_w;k),0)\rightarrow x_{\mathrm{atk}}$ and does not use the clean reference or watermark verifier.}
  \label{fig:pipeline}
\end{figure*}

\subsection{Latent representation and inference path}
The shared encoder and decoder map an image to two full-resolution inputs:
\begin{equation}
  (g,u)=E(x),\qquad \hat{x}=D(g,u),
  \label{eq:encoder-decoder}
\end{equation}
where $g\in[-1,1]^{1\times H\times W}$ is a grayscale structural latent and $u\in[-1,1]^{16\times H\times W}$ is an auxiliary residual latent. For a pair, $(g_c,u_c)=E(x_c)$ and $(g_w,u_w)=E(x_w)$. The channel asymmetry is an inductive bias, not a capacity proof: a full-resolution real-valued $g$ can still encode watermark-relevant information.

The encoder concatenates RGB with a grayscale high-frequency residual, then uses a residual U-Net topology to produce shared full-resolution features~\citep{ronneberger2015unet,he2016resnet}. Two heads map those features to $g$ and $u$. The decoder concatenates $g$, $u$, and a high-frequency residual of $g$ before eight residual blocks and an RGB projection. Both outputs are bounded by $\tanh$. Appendix~\ref{app:architecture} lists every layer.

At inference, the decoder receives
\begin{equation}
  x_{\mathrm{atk}}(k,\alpha)=D(A_g(g_w;k),\alpha u_w).
  \label{eq:latent-attack-output}
\end{equation}
The reported attack sets $\alpha=0$. The point $D(g_w,0)$ isolates auxiliary suppression, and $D(A_g(g_w;k),0)$ adds the deployed structural perturbation. The sweep restores $\alpha u_w$ only to test how this decoder input changes recovery and fidelity.

\subsection{Push and pull supervision}
The two paths receive opposing paired targets:
\begin{align}
  \mathcal{L}_{\mathrm{push}}
    &=\|D(g_w,u_w)-x_w\|_1, \label{eq:push}\\
  \mathcal{L}_{\mathrm{pull}}
    &=\|A_g(g_w;k)-g_c\|_1
      +\lambda_0\|D(A_g(g_w;k),0)-x_c\|_1. \label{eq:pull}
\end{align}
Push requires the joint code to preserve the watermarked image. Pull requires the attacked structural latent and its zero-auxiliary decoding to approach clean targets. Their combination makes $u_w$ an available route for the residual needed by the full reconstruction, while the deployed path is trained without that route. Additional terms preserve the clean reconstruction, discourage unnecessary $u_c$, and constrain perceptual, edge, high-frequency, quantization, and spectral errors. Appendix~\ref{app:full-loss-weights} reports every code-level term and weight.

Paired supervision matters because the pull target preserves the scene, pose, color layout, and texture content of the corresponding watermarked input. The model is not asked to infer a generic clean-image distribution from an unrelated reference. For each pair, it learns which output should be retained by the full path and which output should be approached after the auxiliary intervention. The attack loss never uses decoded bits or detector scores. Its watermark-specific signal is the aligned difference between $x_w$ and $x_c$, together with the two reconstruction targets.

\begin{figure}[t]
  \centering
  \includegraphics[width=0.94\linewidth]{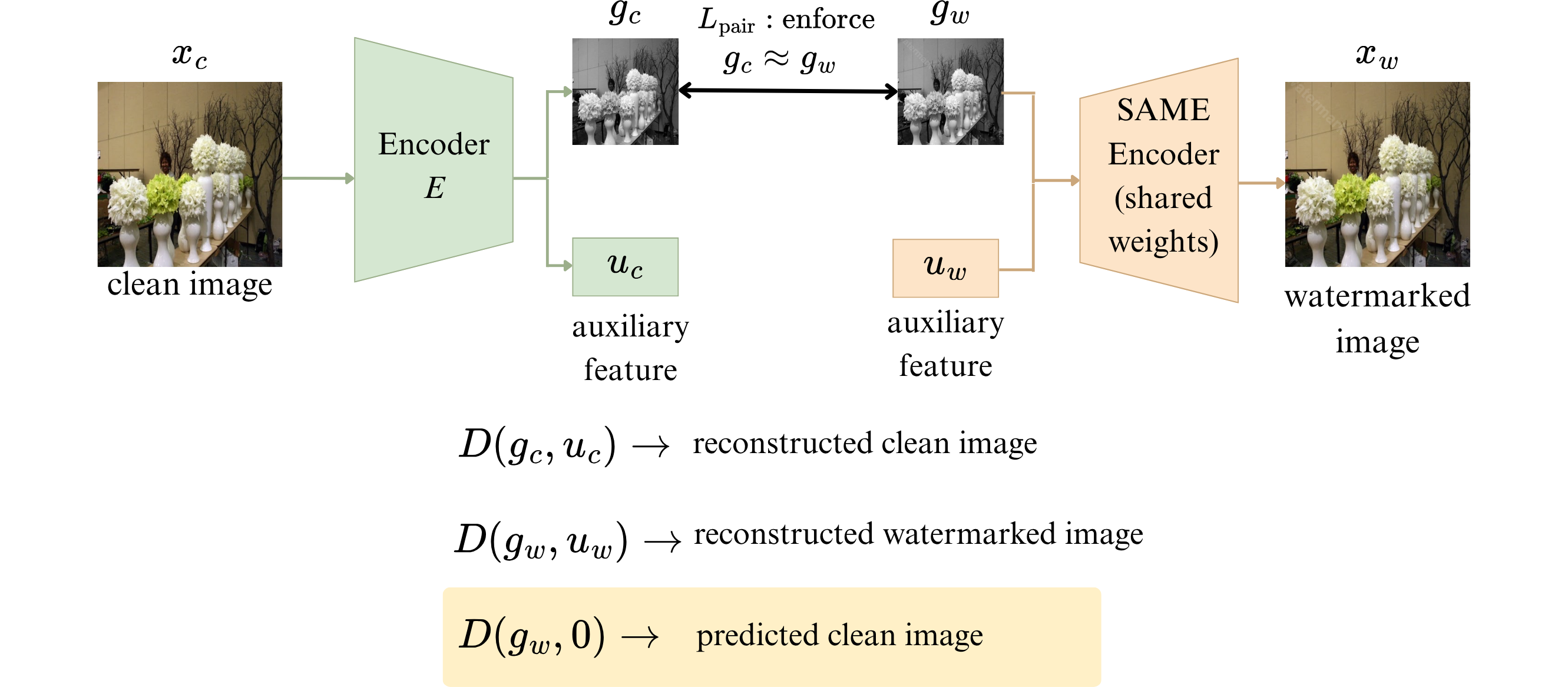}
  \caption{Pair-level supervision. The shared encoder maps aligned clean and watermarked images to two decoder inputs. The full watermarked path is pushed toward $x_w$; the attacked zero-auxiliary path is pulled toward $x_c$.}
  \label{fig:pair-latents}
\end{figure}

The word ``push'' refers to preservation of the watermarked reconstruction, not to increasing watermark strength. The word ``pull'' refers to clean-target supervision on the zero-auxiliary path. In the unperturbed diagnostic, this path is $D(g_w,0)$. The implemented training and deployment path uses $D(A_g(g_w;k),0)$ so that residual suppression and structural attenuation act together. This terminology is kept fixed throughout the experiments.

\begin{table*}[t]
  \caption{Functional groups in the training objective. The first three rows implement the paired push--pull structure; the remaining rows constrain fidelity and optimization.}
  \label{tab:loss-taxonomy}
  \centering
  \small
  \begin{tabularx}{\linewidth}{p{0.19\linewidth}p{0.31\linewidth}X}
    \toprule
    Group & Representative terms & Function \\
    \midrule
    Full-path push & $\|D(g_w,u_w)-x_w\|_1$ & Preserve the watermarked input when both decoder inputs are present. \\
    Structural pull & $\|A_g(g_w;k)-g_c\|_1$ & Align the attacked structural input with the paired clean representation. \\
    Zero-auxiliary pull & $\|D(A_g(g_w;k),0)-x_c\|_1$ & Train the exact form of the deployed decoder input toward the paired clean target. \\
    Reconstruction constraints & Clean reconstruction and $\|u_c\|_1$ & Retain visible content and discourage unnecessary clean residual storage. \\
    Fidelity constraints & Perceptual, high-frequency, and edge losses & Penalize content, texture, and boundary changes in the attacked output. \\
    Spectral constraints & Quantization and multiband magnitude losses & Regularize the attacked structural latent and its frequency profile. \\
    \bottomrule
  \end{tabularx}
\end{table*}

The core objective alone admits degenerate solutions. A decoder could ignore $u$, or the attacked path could approach $x_c$ by blurring details. Clean reconstruction, perceptual features, multiscale high-frequency residuals, and Sobel edges limit these behaviors. The auxiliary floor term prevents early collapse of $u_w$. These terms do not certify a factorization, but they make the zero-auxiliary output an evaluable high-fidelity path rather than an arbitrary branch.

The model is trained for 140 epochs. Stage~1 learns clean and full-path reconstruction. Stage~2 introduces $A_g$ and the zero-auxiliary pull target. Stage~3 activates the complete fidelity and spectral objective. This order prevents the zero-auxiliary output from starting as an unconstrained destructive filter. Architecture blocks and the exact schedule are given in Appendices~\ref{app:architecture} and~\ref{app:training-schedule}.

\subsection{Latent-frequency operator and mechanism test}
The operator applies an rFFT to $g_w$, uses low-, middle-, and high-frequency masks, and reconstructs the perturbed latent through inverse rFFT. For band $B$,
\begin{equation}
  \widehat{A_g(g_w;k)}(\omega)=
  \rho_B\widehat{g_w}(\omega)+\eta_B(\omega),\quad \omega\in B,
  \label{eq:band-attack}
\end{equation}
where $\rho_B$ is a keep ratio and $\eta_B$ is sampled magnitude or phase noise. Stage~3 uses keep ratios $0.90/0.55/0.35$ for low/middle/high bands. A multiband magnitude loss aligns the attacked structural latent with $g_c$. It does not constrain phase or imply pixel equality.

\begin{figure*}[t]
  \centering
  \includegraphics[width=\textwidth]{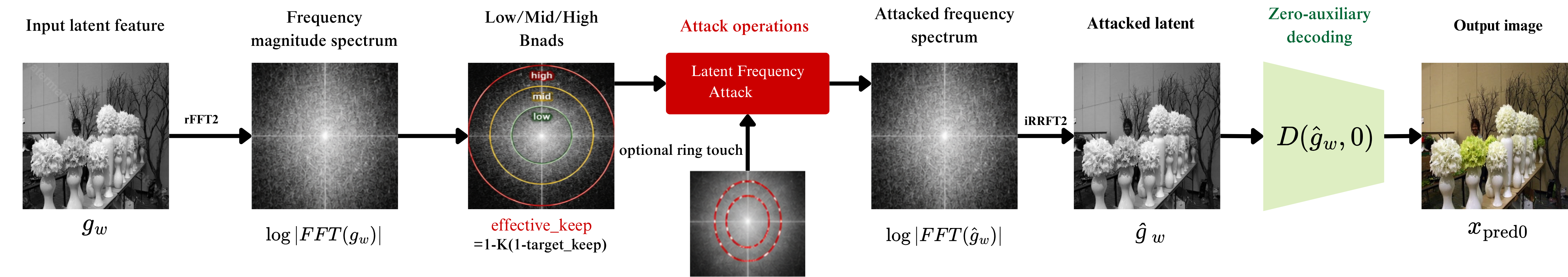}
  \caption{Structural-latent frequency intervention. The rFFT coefficients are divided into radial bands, modified by the $k$-controlled operator, and returned to the decoder through an inverse rFFT.}
  \label{fig:fft-attack}
\end{figure*}

Figure~\ref{fig:fft-attack} separates the learned representation from the attack operator. The Fourier block inside the encoder mixes feature tokens during representation learning. In contrast, $A_g$ is a parameterized intervention on the output $g_w$. The main $(k,u)$ sweep uses band attenuation and magnitude/phase noise. The ring notch for Tree-Ring is disabled in that sweep and enabled only for the radius experiment in Table~\ref{tab:treering-radius-scan}.

The mechanism is tested by holding $A_g(g_w;k)$ fixed and restoring the auxiliary input. Locally,
\begin{equation}
  D(\hat g_w,\alpha u_w)-D(\hat g_w,0)
  =\alpha J_u u_w+R_u(\alpha),
  \label{eq:local-intervention}
\end{equation}
with $\hat g_w=A_g(g_w;k)$. Equation~\eqref{eq:local-intervention} only isolates decoder sensitivity. The watermark-specific evidence comes from the BER response to $\alpha$, not from the expansion itself. The conditional-information interpretation and its unverified premises are deferred to Appendix~\ref{app:formal}.

\section{Experimental Setup}
\label{sec:experiments}

The $(k,u)$ sweep uses DwtDct, DwtDctSvd, RivaGAN, and SSL Watermarking. Evaluation at the selected point adds StegaStamp, Stable Signature, and Tree-Ring. Each method and attack setting uses 100 samples. The coarse grid covers $k,u\in\{0,0.25,0.50,0.75,1.00\}$. The fine grid uses $k\in[0.80,1.20]$ and $u\in\{0,0.03,0.05,0.06,0.10,0.15,0.20\}$. Candidates require average PSNR of at least 25 dB and are ranked by average RR. The same $k=1.10,u=0$ point is then applied across methods; the Tree-Ring radius scan is reported separately in Appendix~\ref{app:extended-results}.

The evaluation addresses four questions. First, does the shared grid contain a point that moves bit recovery toward chance without falling below the fidelity threshold? Second, does restoring $u$ reverse part of that removal effect while the attacked structural input remains fixed? Third, how does the selected point transfer between represented and excluded watermark families, and where does it lie relative to reconstruction and regeneration baselines? Fourth, which objective terms change the observed removal--fidelity profile? The result subsections follow this order so that operating-point selection is separated from mechanism interpretation.

Baselines include JPEG, blur, noise, geometric transforms, BM3D~\citep{dabov2007bm3d}, CtrlRegen~\citep{liu2025ctrlregen}, WatermarkAttacker~\citep{zhao2024provably}, and SADRE variants~\citep{alam2025sadre}. Available outputs are measured with common method-specific evaluators and fidelity metrics. Attack-generation settings and available subsets differ across several baseline rows, so these results are operating-point comparisons rather than an identical-protocol benchmark. Full parameter grids, per-method scores, baseline settings, and complete tables appear in Appendix~\ref{app:extended-results}.

\section{Results}
\label{sec:results}

\subsection{Selected operating point}
The fine sweep selects $k=1.10,u=0$. Its average BER is $0.3958$, closer to $0.5$ than the other reported shared settings, while PSNR remains $31.07$ dB. Table~\ref{tab:main-result} also reports nearby points to show the measured local tradeoff. We treat $k=1.10$ as the selected grid point, not as evidence of a continuous or seed-stable optimum.

\begin{table}[!t]
  \caption{Shared fine-sweep settings over DwtDct, DwtDctSvd, RivaGAN, and SSL Watermarking. Bold BER is closest to $0.5$; bold PSNR/SSIM is larger.}
  \label{tab:main-result}
  \centering
  \footnotesize
  \begin{tabular}{ccccc}
    \toprule
    $k$ & $u$ & BER $\rightarrow0.5$ & PSNR$\uparrow$ & SSIM$\uparrow$ \\
    \midrule
    1.10 & 0.00 & \textbf{0.3958} & 31.07 & 0.9554 \\
    1.00 & 0.00 & 0.3938 & 31.07 & 0.9553 \\
    0.90 & 0.00 & 0.3867 & 31.11 & 0.9586 \\
    0.90 & 0.03 & 0.3720 & \textbf{31.30} & \textbf{0.9591} \\
    \bottomrule
  \end{tabular}
\end{table}

The two controls play different roles. With $u=0$, changing $k$ from $0.90$ to $1.10$ moves BER from $0.3867$ to $0.3958$ with little change in fidelity. Restoring $u=0.03$ at $k=0.90$ raises PSNR by $0.19$ dB but moves BER to $0.3720$. The full coarse and fine tables in Appendix~\ref{app:extended-results} retain the points omitted here.

\subsection{Auxiliary-input intervention}
Table~\ref{tab:u-sensitivity-main} groups the fine sweep by residual scale. The change from $u=0$ to $u=0.03$ is small in fidelity: PSNR rises by $0.06$ dB and SSIM rises by $0.0003$. BER nevertheless moves from $0.3893$ to $0.3695$, farther from random guessing. At $u=0.15$, BER reaches $0.3357$ and PSNR falls to $28.23$ dB. Restoring $u$ therefore does not act as a uniformly beneficial detail correction after $g$ has been attacked.

\begin{table}[!t]
  \caption{Fine-sweep averages by auxiliary scale. Bold BER is closest to $0.5$; bold PSNR/SSIM is larger.}
  \label{tab:u-sensitivity-main}
  \centering
  \footnotesize
  \begin{tabular}{cccc}
    \toprule
    $u$ & BER $\rightarrow0.5$ & PSNR$\uparrow$ & SSIM$\uparrow$ \\
    \midrule
    0.00 & \textbf{0.3893} & 30.99 & 0.9568 \\
    0.03 & 0.3695 & \textbf{31.05} & \textbf{0.9571} \\
    0.06 & 0.3521 & 30.83 & \textbf{0.9571} \\
    0.15 & 0.3357 & 28.23 & 0.9548 \\
    0.20 & 0.3426 & 26.09 & 0.9509 \\
    \bottomrule
  \end{tabular}
\end{table}

This intervention supports a functional statement: the evaluated decoders recover more payload information when the auxiliary input is restored. It does not identify which pixels or frequencies carry that information, and it does not show that $g$ is information-free. The conclusion is limited to branch dependence under the trained model and tested watermark families.

\subsection{Seen and unseen families}
At $k=1.10,u=0$, DwtDctSvd reaches BER $0.4856$, RivaGAN $0.3697$, and StegaStamp $0.4251$ among the seen bit-decoder families. The unseen DwtDct, SSL Watermarking, and Stable Signature families reach $0.4106$, $0.3207$, and $0.2275$. The spread prevents a single generalization claim. It shows transfer to excluded families with method-dependent strength. Tree-Ring remains separate because its detector output is not a bit payload; its Acc Removal is $0.3200$ at PSNR $26.81$ dB.

\begin{table}[!t]
  \caption{Selected-point results on all methods. Tree-Ring reports Acc Removal; other rows report BER, with values closer to $0.5$ preferred.}
  \label{tab:all-methods-main}
  \centering
  \footnotesize
  \begin{tabular}{lcccc}
    \toprule
    Method & Family & BER/Acc Rem. & PSNR & SSIM \\
    \midrule
    DwtDct & Unseen & 0.4106 & 31.69 & 0.9656 \\
    DwtDctSvd & Seen & 0.4856 & 31.54 & 0.9659 \\
    RivaGAN & Seen & 0.3697 & 31.65 & 0.9597 \\
    SSL Watermarking & Unseen & 0.3207 & 29.43 & 0.9303 \\
    StegaStamp & Seen & 0.4251 & 31.37 & 0.9470 \\
    Stable Signature & Unseen & 0.2275 & 28.35 & 0.8934 \\
    Tree-Ring & Seen & 0.3200 & 26.81 & 0.8799 \\
    \bottomrule
  \end{tabular}
\end{table}

The two transform-domain rows give the closest BERs to chance, despite DwtDct being absent from paired training. RivaGAN and StegaStamp show intermediate removal at PSNR above 31 dB. Transfer is weaker for SSL Watermarking and Stable Signature. The result is compatible with family-level transfer but also shows that a shared $(k,u)$ point does not equalize difficulty across watermark designs.

\subsection{Removal and fidelity relative to baselines}
Table~\ref{tab:baseline-summary} summarizes six bit-decoder methods with mean RR, PSNR, and SSIM. Tree-Ring is listed in separate columns. CtrlRegen and WatermarkAttacker obtain higher mean RR than our selected point, but at 21.28 and 24.13 dB PSNR. SADRE-Cheng reaches mean RR $0.7692$ at similar mean PSNR, while our point gives RR $0.7464$ and higher mean SSIM ($0.9436$ versus $0.8792$). On Tree-Ring, our point has the largest Acc Removal in this subset. These aggregates describe a measured fidelity--removal profile; they do not establish uniform dominance.

\begin{table*}[t]
  \caption{Compact baseline comparison. Bit metrics average DwtDct, DwtDctSvd, RivaGAN, SSL Watermarking, StegaStamp, and Stable Signature. Tree-Ring detector results are not included in those averages.}
  \label{tab:baseline-summary}
  \centering
  \footnotesize
  \resizebox{\linewidth}{!}{%
  \begin{tabular}{lcccccc}
    \toprule
    Attack & Mean RR$\uparrow$ & Mean PSNR$\uparrow$ & Mean SSIM$\uparrow$ & Tree-Ring Acc Removal$\uparrow$ & Tree-Ring PSNR$\uparrow$ & Tree-Ring SSIM$\uparrow$ \\
    \midrule
    JPEG & 0.1774 & \textbf{35.87} & \textbf{0.9669} & 0.0000 & \textbf{43.18} & \textbf{0.9923} \\
    Gaussian noise & 0.4769 & 20.99 & 0.4947 & 0.0100 & 22.45 & 0.4555 \\
    CtrlRegen & \textbf{0.9276} & 21.28 & 0.6139 & 0.2200 & 21.68 & 0.6570 \\
    WatermarkAttacker & 0.8251 & 24.13 & 0.7049 & 0.1500 & 24.98 & 0.7702 \\
    SADRE-Cheng & 0.7692 & 30.73 & 0.8792 & 0.0400 & 31.36 & 0.8812 \\
    Ours ($k=1.10,u=0$) & 0.7464 & 30.67 & 0.9436 & \textbf{0.3200} & 26.81 & 0.8799 \\
    \bottomrule
  \end{tabular}}
\end{table*}

The full baseline tables show why raw BER or TPR cannot be read alone. Brightness, rotation, and cropping remove some watermarks at PSNR values near 7--10 dB. JPEG preserves quality but leaves most learned payloads recoverable. Regeneration attacks move more decoders toward chance, although their fidelity is lower. Our selected point is strongest where the paired projection can suppress recoverable evidence without broad image reconstruction; Stable Signature remains a weak case.

\begin{figure*}[t]
  \centering
  \includegraphics[width=0.72\textwidth,height=0.70\textheight,keepaspectratio]{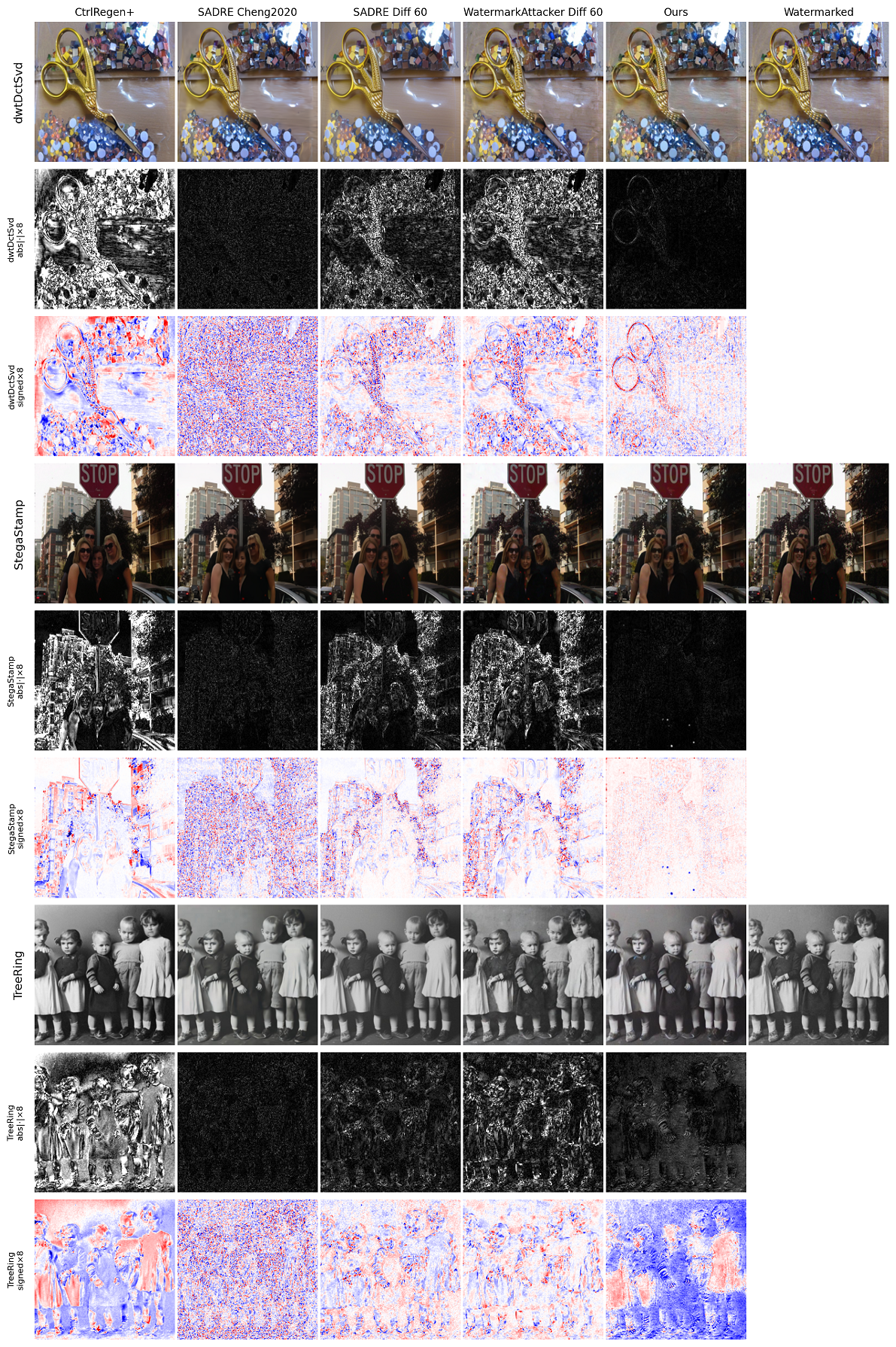}
  \caption{Representative outputs, absolute residuals $|x_{\mathrm{atk}}-x_w|\times8$, and signed residuals $(x_{\mathrm{atk}}-x_w)\times8$. Red and blue denote positive and negative signed changes. The full grid is in Appendix~\ref{app:full-double-residual}.}
  \label{fig:double-residual-sample}
\end{figure*}

Figure~\ref{fig:double-residual-sample} compares three representative outputs. The absolute residual records where each attack changes $x_w$, while the signed residual records the direction of change. Regeneration and diffusion methods produce broad changes over object boundaries and textures. Our output has a weaker residual on DwtDctSvd and StegaStamp. Tree-Ring produces a broader footprint, consistent with the lower fidelity in Table~\ref{tab:all-methods-main}. These images are qualitative diagnostics; they do not measure watermark information or prove frequency localization.

Tree-Ring uses a structured frequency detector, so its ring radius is a detector-specific control rather than a bit-decoder parameter. Table~\ref{tab:treering-radius-scan} shows that a ring notch at $r=0.50$ increases Acc Removal from $0.3200$ to $0.3900$. PSNR decreases from $26.81$ to $25.47$ dB. Larger radii recover fidelity but lose the removal gain. This tradeoff is reported separately from all BER averages.

\begin{table}[!t]
  \caption{Tree-Ring ring-notch radius scan at $k=1.10,u=0$.}
  \label{tab:treering-radius-scan}
  \centering
  \footnotesize
  \begin{tabular}{lccc}
    \toprule
    Attack & Acc Removal$\uparrow$ & PSNR$\uparrow$ & SSIM$\uparrow$ \\
    \midrule
    No ring notch & 0.3200 & \textbf{26.81} & \textbf{0.8799} \\
    $r=0.40$ & 0.3400 & 26.60 & 0.8723 \\
    $r=0.45$ & 0.3600 & 26.31 & 0.8727 \\
    $r=0.50$ & \textbf{0.3900} & 25.47 & 0.8724 \\
    $r=0.55$ & 0.3500 & 25.40 & 0.8739 \\
    $r=0.60$ & 0.3200 & 26.24 & 0.8770 \\
    $r=0.65$ & 0.3200 & 26.71 & 0.8792 \\
    \bottomrule
  \end{tabular}
\end{table}

\subsection{Ablation}
\label{sec:ablation}
Six retrained variants are evaluated at $k=1.10,u=0$. Table~\ref{tab:ablation-mean} averages the six bit-decoder methods and excludes Tree-Ring. Removing the multiband magnitude loss changes mean BER by $-0.044$, away from chance, while PSNR increases by $1.19$ dB. In these runs the spectral term is associated with a stronger attack at a fidelity cost. Removing the scheduled drop0 term changes mean BER by $+0.038$ and PSNR by $-3.37$ dB.

\begin{table}[!t]
  \caption{Mean ablation changes over six bit-decoder methods. Tree-Ring is excluded. Positive $\Delta$BER moves toward $0.5$ because all values remain below $0.5$.}
  \label{tab:ablation-mean}
  \centering
  \footnotesize
  \begin{tabularx}{\linewidth}{>{\raggedright\arraybackslash}Xcc}
    \toprule
    Variant & $\Delta$BER & $\Delta$PSNR \\
    \midrule
    w/o spectral alignment & $-0.044$ & $+1.19$ \\
    w/o scheduled drop0 term & $+0.038$ & $-3.37$ \\
    w/o perceptual constraint & $+0.010$ & $-0.31$ \\
    w/o high-frequency residual & $-0.001$ & $-0.01$ \\
    Fourier groups 4 & $+0.004$ & $-0.06$ \\
    Fourier groups 16 & $+0.001$ & $-0.06$ \\
    \bottomrule
  \end{tabularx}
\end{table}

The drop0 result is not a clean ablation of zero-auxiliary supervision. In Stage~3, $\mathcal{L}_{\mathrm{drop0}}$ and $\mathcal{L}_{\mathrm{recov}}$ are the same pixel residual with effective combined weight 16. The variant removes only the scheduled term; $\mathcal{L}_{\mathrm{recov}}$ remains. The two perceptual names also reuse one pooled-VGG value with effective weight 10. These duplicate terms make the current variants schedule and weight ablations. Appendix~\ref{app:full-loss-weights} documents the implemented objective, and Appendix~\ref{app:full-ablation} reports every method--variant pair.

The per-method rows clarify the aggregate changes. Without spectral alignment, removal weakens on six of seven reported methods even though PSNR rises. This is an association within the trained variants, not proof that the magnitude term alone causes removal. Removing the scheduled drop0 term gives BERs closer to $0.5$ on several bit decoders but lowers PSNR by about 3 dB. Tree-Ring Acc Removal also falls from $0.32$ to $0.17$. The remaining clean-target residual means the result concerns Stage~2 exposure and the effective Stage~3 weight.

The perceptual ablation has a smaller mean effect and a heterogeneous method response. Stable Signature BER moves from $0.2275$ to $0.3863$, but SSIM falls from $0.8934$ to $0.8360$. RivaGAN, SSL Watermarking, DwtDctSvd, and StegaStamp move farther from chance without this term. The high-frequency residual and Fourier group-count variants change mean BER and PSNR by less than $0.01$. Repeated seeds are required before treating these small differences as resolved component effects.

\section{Discussion}

The evidence chain is specific. Paired supervision supplies an image-aligned clean target. Push keeps $D(g_w,u_w)$ faithful to $x_w$, whereas pull trains $D(A_g(g_w;k),0)$ toward $x_c$. Zeroing $u$ then selects the trained pull path, and the $u$-restoration experiment measures how much decoder recovery returns. This sequence links the training resource, intervention, and observable outcome. It also separates the contribution from prior paired VAE removal: the empirical question is how recovery changes across two controlled decoder inputs.

The selected $k=1.10$ point controls the structural intervention, but the auxiliary scale is the cleaner mechanism variable. Across $u=0$ settings, the reported changes in $k$ alter average BER by less than $0.01$ and leave PSNR near 31 dB. Restoring a small fraction of $u_w$ changes BER more sharply. The experiment therefore supports the claim that the decoder response depends on the auxiliary input. It does not determine whether that response is caused by payload-specific evidence, watermark-correlated texture, or another residual feature.

The baseline comparison also separates attack success from visible damage. CtrlRegen has the highest mean RR in Table~\ref{tab:baseline-summary}, but its mean PSNR is 21.28 dB. SADRE-Cheng and our attack have similar mean PSNR, while our SSIM is higher and SADRE-Cheng has slightly higher mean RR. These tradeoffs do not support a claim of universal superiority. They show that paired push--pull learning occupies a distinct operating region from aggressive regeneration and high-fidelity but weak distortions.

The comparison suggests a defense criterion. Watermark recovery should depend on image structure that cannot be removed by a learned content-preserving projection. This criterion is stronger than survival under a fixed list of distortions. It should be tested with paired learned attackers and reported under explicit fidelity constraints. Detector-based methods still require calibrated detector metrics and method-specific tuning; the Tree-Ring radius scan is therefore kept outside bit-decoder summaries.

The two zero-auxiliary expressions serve different experimental roles. $D(g_w,0)$ is the direct branch intervention: it asks what the decoder produces when the auxiliary input is removed and the structural latent is unchanged. $D(A_g(g_w;k),0)$ is the complete attack: it combines the same branch intervention with a controlled perturbation of the remaining structural latent. Treating both outputs as evidence for one undifferentiated mechanism would conflate residual suppression with frequency attenuation. Our parameter sweep separates them through $k$, and the ablation separates the learned spectral constraint from the zero-auxiliary schedule. A stronger causal study would report all four counterfactuals, $D(g_w,u_w)$, $D(g_w,0)$, $D(A_g(g_w;k),u_w)$, and $D(A_g(g_w;k),0)$, for every watermark family.

Paired supervision also changes how the attack should be described. The deployed mapping is query-free, but the training resource is stronger than an unpaired image collection. The reported operating point is selected after evaluator-side decoding. The relevant access axes are therefore separate: paired data are available during training, the watermark verifier is available during offline evaluation, and neither is supplied when one image is attacked. This distinction avoids calling the entire workflow ``decoder-free.'' It also permits a direct comparison with the ICLR 2025 workshop solution. Both settings use paired supervision and single-image deployment. The proposed method adds a two-input decoder intervention and a latent-frequency structural path.

The family split gives a limited transfer test rather than a universal no-box result. DwtDct is excluded from paired training but is structurally related to the represented DwtDctSvd family. SSL Watermarking and Stable Signature are also excluded, yet their recovery interfaces and embedding mechanisms differ from those in training. Their lower removal scores show that exclusion alone does not define difficulty. Future evaluation should vary family, implementation, payload, key, and image distribution independently. This would reveal whether transfer comes from shared signal structure, from the paired image prior, or from incidental overlap between implementations.

\section{Limitations}

The four-method parameter sweep and seven-method selected-point evaluation use 100 samples per method. They do not cover broader datasets, implementations, resolutions, key distributions, or payload lengths. The finite $(k,u)$ grid supports the selected point $k=1.10,u=0$ but cannot establish a continuous optimum or repeatability across seeds. Nearby grid points show similar aggregate values, yet one run cannot distinguish a smooth response region from sampling variation. Repeated training seeds and per-image confidence intervals are required before making a stability claim.

Several baseline outputs come from different attack-generation settings and available subsets. We apply common method-specific evaluators and fidelity calculations to those outputs, but this metric alignment does not make the generation protocol identical. Runtime, compute budget, input resolution, stochastic sampling, and model-specific preprocessing may differ. Table~\ref{tab:baseline-summary} should therefore be read as an operating-point comparison. A controlled benchmark would regenerate every attack from the same image manifest, preserve per-image pairing, report failure counts, and tune each baseline under the same quality constraint.

PSNR and SSIM are measured against the watermarked input and do not fully capture semantic or perceptual preservation. This limitation is most visible for regeneration methods, which may retain recognizable content while changing fine texture, and for geometric attacks, which can obtain detector failure after large pixel displacement. LPIPS or another learned perceptual metric would add a complementary view~\citep{zhang2018lpips}. Human inspection remains useful for semantic drift, but it should follow a defined sampling and rating protocol rather than a few selected examples.

The mechanism theory is post-hoc. Mutual information is not estimated, the one-channel $g$ is not a proven bottleneck, and the multiband magnitude loss does not constrain Fourier phase. The 98.75\% $u$-probe accuracy is preliminary because it comes from one train/test split. The 21.25\% $g$-probe accuracy cannot be interpreted as absence of information: reversing labels, changing the split, or using calibrated scores may expose discrimination. A confirmatory analysis should report AUROC, repeated cross-validation, uncertainty, and label-permutation tests. It should also vary the $g$ channel count and quantization strength before attributing the response to a capacity bottleneck.

The available ablations contain one summarized run per variant and no uncertainty estimates. Their bookkeeping also duplicates two Stage~3 residuals. A reproducible release should include the exact data manifest, family split, random seeds, configuration files, evaluation commands, per-image scores, and checkpoints. Future ablations should consolidate the duplicate pixel and perceptual terms before retraining. This would turn the current schedule/weight comparisons into tests of the intended objective components.

\section{Conclusion}

We studied watermark removal with paired clean--watermarked training and single-image inference. The push path reconstructs the watermarked image; the pull path trains a zero-auxiliary decoding toward the paired clean target. At $k=1.10,u=0$, the attack reaches average BER $0.3958$ at PSNR $31.07$ dB in the four-method sweep. Restoring $u$ moves recovery away from chance, which supports decoder dependence on the auxiliary input in this model. The result motivates paired learned attacks as part of fidelity-constrained watermark evaluation, while the full mechanism claim remains conditional on stronger representation tests.

\appendix

\section{Formal Objective and Post-Hoc Mechanism Account}
\label{app:formal}

\subsection{Fidelity-constrained removal risk}
Let $\mathcal{W}$ be a watermarking scheme and let $S_{\mathcal{W}}(y,m)\in[0,1]$ measure recoverability of message $m$ from attacked image $y$. For bit decoders,
\begin{equation}
  S_{\mathcal{W}}(y,m)=2\left|\mathrm{BER}_{\mathcal{W}}(y,m)-\tfrac{1}{2}\right|.
  \label{eq:app-recoverability}
\end{equation}
For an attack class $\mathcal{A}$ and visible-distortion budget $\epsilon$, define
\begin{equation}
\begin{aligned}
  \mathcal{R}_{\epsilon}(\mathcal{W},\mathcal{A})
  &=\sup_{A\in\mathcal{A}}\mathbb{E}_{(x_w,m)}
    [1-S_{\mathcal{W}}(A(x_w),m)]\\
  &\mathrm{s.t.}\quad
  \mathbb{E}_{x_w}[\ell_{\mathrm{vis}}(A(x_w),x_w)]\le\epsilon .
\end{aligned}
  \label{eq:app-fidelity-risk}
\end{equation}
This quantity separates image destruction from removal under a stated fidelity budget. It also makes the attack class explicit. Robustness against a hand-designed distortion subset does not bound the risk over a larger class that contains paired learned attacks~\citep{goodfellow2015explaining,madry2018towards}.

\subsection{Conditional-information interpretation}
For random payload $m$, the chain rule gives
\begin{equation}
  I(m;(g_w,u_w))=I(m;g_w)+I(m;u_w\mid g_w).
  \label{eq:app-chain-rule}
\end{equation}
The identity partitions information in the joint representation but does not determine either term. An auxiliary-concentration interpretation would additionally require two premises: the full path retains payload-relevant information, and the attacked structural path contains little of it. One possible statement is
\begin{equation}
\begin{aligned}
 I(m;(g_w,u_w))&\ge I(m;x_w)-\gamma(\delta),\\
 I(m;A_g(g_w;k))&\le\varepsilon_I,
 \qquad \delta=\mathbb{E}\|D(g_w,u_w)-x_w\|.
\end{aligned}
\label{eq:app-information-premises}
\end{equation}
Data processing would then give
\begin{equation}
  I(m;D(A_g(g_w;k),0))\le I(m;A_g(g_w;k)).
  \label{eq:app-data-processing}
\end{equation}
The present training objective does not optimize mutual information, and the experiment does not estimate $\varepsilon_I$. Equations~\eqref{eq:app-chain-rule}--\eqref{eq:app-data-processing} are therefore a conditional, post-hoc explanation. The $u$ intervention measures decoder dependence but cannot establish exclusive storage or statistical disentanglement~\citep{belghazi2018mine,poole2019variational}.

\subsection{Constrained push--pull view}
The main reconstruction targets can be summarized as
\begin{equation}
\begin{aligned}
 \min_{E,D}\quad &\mathbb{E}\|D(A_g(g_w;k),0)-x_c\|_2^2\\
 \mathrm{s.t.}\quad
 &\mathbb{E}\|A_g(g_w;k)-g_c\|_2\le\varepsilon_g,\\
 &\mathbb{E}\|D(g_w,u_w)-x_w\|_2\le\delta.
\end{aligned}
\label{eq:app-constrained-push-pull}
\end{equation}
Using $u_w$ is one way for the optimizer to satisfy the full-path constraint while the zero-auxiliary output follows the clean target. It is not the unique solution. A sufficiently expressive encoder can retain evidence in $g$, distribute it across both branches, or exploit decoder nonlinearities. The theory becomes predictive only after its representation premises are independently tested.

The empirical checks suggested by this account are: (i) restoring $u$ should move BER away from $0.5$ when visible fidelity remains high; (ii) removing spectral alignment should reduce RR if the attacked structural latent no longer matches the clean target; and (iii) removing zero-auxiliary supervision should reduce the fidelity of the deployed path. The current residual-scale experiment addresses the first check. The second is descriptive because only one summarized run is available per ablation. The third is unresolved because the drop0 ablation leaves a duplicate Stage~3 clean-target residual active.

\section{Extended Parameter Search and Per-Method Results}
\label{app:extended-results}

\subsection{Coarse and fine grids}
Table~\ref{tab:coarse-corners} reports the four diagnostic corners. $K0U0$ has average BER closest to $0.5$ but only 22.51 dB PSNR. Restoring the full auxiliary input at $K0U1$ raises PSNR to 37.80 dB and makes the payload recoverable. $K1U0$ is the only corner that combines high removal with PSNR above 30 dB.

\begin{table}[!t]
  \caption{Coarse-grid corners. Method BERs follow SSL Watermarking/DwtDct/DwtDctSvd/RivaGAN. Bold BER is closest to $0.5$; bold PSNR/SSIM is larger.}
  \label{tab:coarse-corners}
  \centering
  \scriptsize
  \resizebox{\linewidth}{!}{%
  \begin{tabular}{lccccc l}
    \toprule
    Setting & $k$ & $u$ & Avg. BER $\rightarrow0.5$ & PSNR$\uparrow$ & SSIM$\uparrow$ & Method BERs \\
    \midrule
    $K0U0$ & 0.00 & 0.00 & \textbf{0.4042} & 22.51 & 0.9144 & 0.339/0.444/0.489/0.344 \\
    $K1U0$ & 1.00 & 0.00 & 0.3913 & 31.07 & 0.9553 & 0.316/0.414/0.470/0.365 \\
    $K0U1$ & 0.00 & 1.00 & 0.0697 & \textbf{37.80} & \textbf{0.9883} & 0.012/0.240/0.026/0.001 \\
    $K1U1$ & 1.00 & 1.00 & 0.1137 & 27.95 & 0.8583 & 0.125/0.225/0.105/0.000 \\
    \bottomrule
  \end{tabular}}
\end{table}

\begin{table}[!t]
  \caption{Full set of selected fine-sweep points. Method BERs follow SSL Watermarking/DwtDct/DwtDctSvd/RivaGAN.}
  \label{tab:main-result-full}
  \centering
  \scriptsize
  \resizebox{\linewidth}{!}{%
  \begin{tabular}{lccccc l}
    \toprule
    Setting & $k$ & $u$ & Avg. BER $\rightarrow0.5$ & PSNR$\uparrow$ & SSIM$\uparrow$ & Method BERs \\
    \midrule
    Attack optimum & 1.10 & 0.00 & \textbf{0.3958} & 31.07 & 0.9554 & 0.324/0.418/0.479/0.362 \\
    Nearby point & 1.00 & 0.00 & 0.3938 & 31.07 & 0.9553 & 0.324/0.411/0.473/0.368 \\
    Balanced point & 0.90 & 0.00 & 0.3867 & 31.11 & 0.9586 & 0.319/0.418/0.463/0.348 \\
    Nonzero-$u$ point & 0.90 & 0.03 & 0.3720 & \textbf{31.30} & 0.9591 & 0.311/0.414/0.486/0.277 \\
    Higher-$u$ point & 0.90 & 0.05 & 0.3550 & 31.29 & \textbf{0.9593} & 0.312/0.413/0.475/0.219 \\
    \bottomrule
  \end{tabular}}
\end{table}

\begin{table}[!t]
  \caption{Fine-sweep averages grouped by auxiliary scale. Deltas are relative to $u=0$.}
  \label{tab:u-sensitivity}
  \centering
  \small
  \resizebox{\linewidth}{!}{%
  \begin{tabular}{cccccc}
    \toprule
    $u$ & BER $\rightarrow0.5$ & $\Delta$BER & PSNR$\uparrow$ & $\Delta$PSNR & SSIM$\uparrow$ \\
    \midrule
    0.00 & \textbf{0.3893} & 0.0000 & 30.99 & 0.00 & 0.9568 \\
    0.03 & 0.3695 & $-0.0198$ & \textbf{31.05} & $+0.06$ & \textbf{0.9571} \\
    0.05 & 0.3570 & $-0.0323$ & 30.94 & $-0.05$ & \textbf{0.9571} \\
    0.06 & 0.3521 & $-0.0372$ & 30.83 & $-0.16$ & \textbf{0.9571} \\
    0.10 & 0.3399 & $-0.0494$ & 30.01 & $-0.98$ & 0.9566 \\
    0.15 & 0.3357 & $-0.0536$ & 28.23 & $-2.76$ & 0.9548 \\
    0.20 & 0.3426 & $-0.0467$ & 26.09 & $-4.90$ & 0.9509 \\
    \bottomrule
  \end{tabular}}
\end{table}

\subsection{Per-method results}
Table~\ref{tab:method-main} compares the selected point with a quality-oriented nonzero-$u$ setting. Restoring $u=0.03$ has the largest decoder effect on RivaGAN, whose BER moves from $0.3619$ to $0.2772$.

\begin{table*}[t]
  \caption{Per-method fine-sweep metrics. Bold BER is closest to $0.5$; bold PSNR/SSIM is larger.}
  \label{tab:method-main}
  \centering
  \scriptsize
  \resizebox{\linewidth}{!}{%
  \begin{tabular}{lcccccccccccc}
    \toprule
    Setting & \multicolumn{3}{c}{DwtDct} & \multicolumn{3}{c}{DwtDctSvd} & \multicolumn{3}{c}{RivaGAN} & \multicolumn{3}{c}{SSL Watermarking} \\
    \cmidrule(lr){2-4}\cmidrule(lr){5-7}\cmidrule(lr){8-10}\cmidrule(lr){11-13}
    & BER$\rightarrow0.5$ & PSNR & SSIM & BER$\rightarrow0.5$ & PSNR & SSIM & BER$\rightarrow0.5$ & PSNR & SSIM & BER$\rightarrow0.5$ & PSNR & SSIM \\
    \midrule
    $k=1.10,u=0$ & \textbf{0.4181} & 31.68 & 0.9656 & 0.4794 & 31.54 & 0.9659 & \textbf{0.3619} & 31.65 & 0.9597 & \textbf{0.3237} & 29.42 & 0.9302 \\
    $k=0.90,u=0.03$ & 0.4141 & \textbf{31.85} & \textbf{0.9683} & \textbf{0.4859} & \textbf{31.61} & \textbf{0.9685} & 0.2772 & \textbf{31.83} & \textbf{0.9627} & 0.3110 & \textbf{29.92} & \textbf{0.9368} \\
    \bottomrule
  \end{tabular}}
\end{table*}

\begin{table}[!t]
  \caption{Per-method points recomputed from the scan result files. Strongest attack maximizes RR with PSNR $\ge25$ dB; constrained rows maximize PSNR subject to an RR floor.}
  \label{tab:ku-scan-method-optima}
  \centering
  \scriptsize
  \resizebox{\linewidth}{!}{%
  \begin{tabular}{llrrrrr}
    \toprule
    Method & Selection & $k$ & $u$ & BER$\rightarrow0.5$ & PSNR$\uparrow$ & SSIM$\uparrow$ \\
    \midrule
    DwtDct & Strongest attack & 0.8500 & 0.2500 & \textbf{0.5003} & 25.47 & 0.9613 \\
    DwtDct & Best recovery & 0.0000 & 1.0000 & 0.2403 & \textbf{37.27} & 0.9900 \\
    DwtDct & RR $\ge0.50$ best PSNR & 0.9250 & 0.0375 & 0.4153 & \textbf{31.88} & 0.9679 \\
    \midrule
    DwtDctSvd & Strongest attack & 0.5000 & 0.2500 & \textbf{0.5000} & 27.02 & 0.9661 \\
    DwtDctSvd & Best recovery & 0.0000 & 1.0000 & 0.0262 & \textbf{37.59} & 0.9896 \\
    DwtDctSvd & RR $\ge0.50$ best PSNR & 0.9400 & 0.0163 & 0.4675 & \textbf{31.67} & 0.9677 \\
    \midrule
    RivaGAN & Strongest attack & 1.0000 & 0.0000 & \textbf{0.3694} & 31.66 & 0.9597 \\
    RivaGAN & Best recovery & 0.0000 & 1.0000 & 0.0006 & \textbf{38.63} & 0.9871 \\
    RivaGAN & RR $\ge0.60$ best PSNR & 0.9431 & 0.0050 & 0.3147 & \textbf{31.82} & 0.9616 \\
    \midrule
    SSL Watermarking & Strongest attack & 1.0000 & 0.2000 & \textbf{0.3570} & 25.37 & 0.9258 \\
    SSL Watermarking & Best recovery & 0.0000 & 1.0000 & 0.0115 & \textbf{37.70} & 0.9865 \\
    SSL Watermarking & RR $\ge0.60$ best PSNR & 0.8000 & 0.0500 & 0.3042 & \textbf{30.04} & 0.9416 \\
    \bottomrule
  \end{tabular}}
\end{table}

\begin{figure*}[t]
  \centering
  \includegraphics[width=0.48\textwidth]{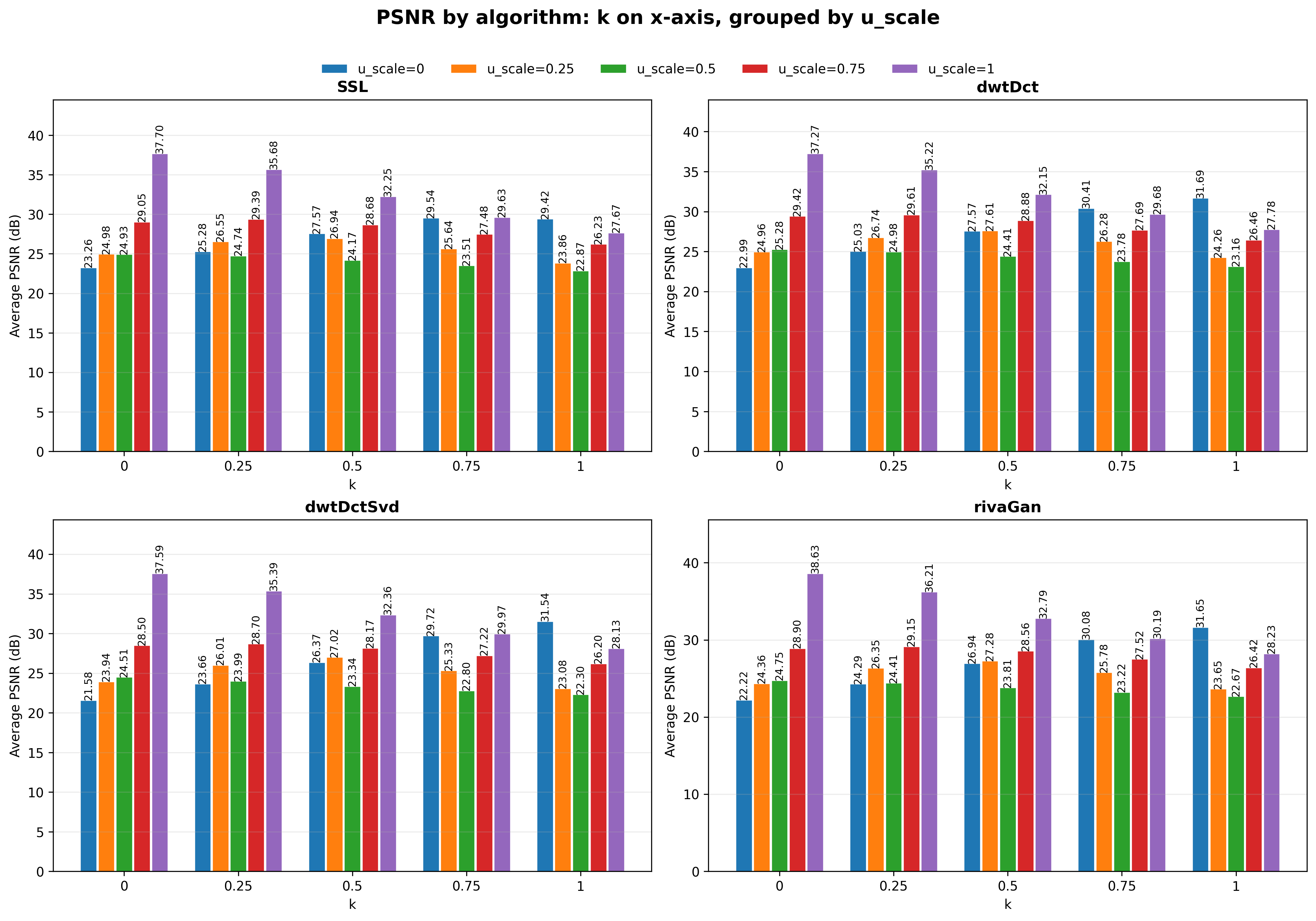}\hfill
  \includegraphics[width=0.48\textwidth]{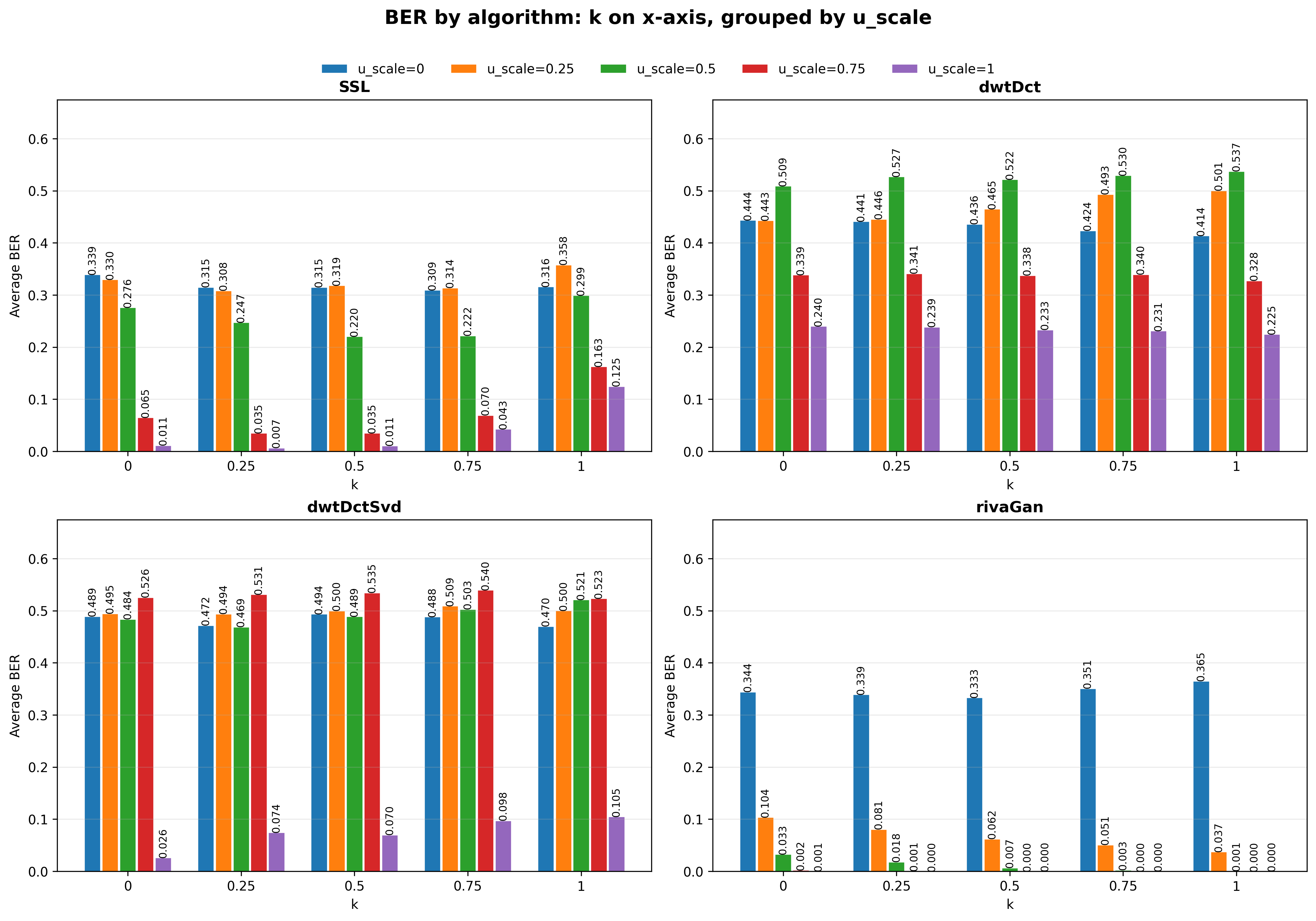}
  \caption{Full PSNR (left) and BER (right) views across attack strength $k$ and auxiliary scale $u$. BER should be read by proximity to $0.5$.}
  \label{fig:app-ku-bars}
\end{figure*}

\subsection{Tree-Ring radius and representation probes}
The ring-notch option is disabled for the reported $(k,u)$ sweep and module ablations. It is enabled only for the detector-specific scan in Table~\ref{tab:treering-radius-scan} in the main paper.

On 399 matched representations, one train/test linear probe distinguishes watermarked from attacked samples with 98.75\% accuracy using $u$. This is preliminary evidence of a linearly accessible intervention effect. The corresponding $g$ probe gives 21.25\% accuracy against a nominal 50\% chance level. A below-chance value cannot be read as ``no information'' because reversed labels, split instability, or calibration failure may still yield discriminability. A confirmatory probe should report score-based AUROC, repeated cross-validation, confidence intervals, and label-permutation tests.

\clearpage
\section{Complete Baseline Comparisons}
\label{app:baseline-comparisons}

The tables below retain the method-level results used to derive Table~\ref{tab:baseline-summary}. All available outputs are evaluated with method-specific decoders or detectors and common fidelity metrics. The attack-generation configurations and result subsets are not identical across rows.

\begin{table*}[!b]
  \caption{Conventional image-space attacks and our selected point. Bold BER is closest to $0.5$; bold detector removal and fidelity metrics are larger.}
  \label{tab:traditional-attack-comparison}
  \centering
  \scriptsize
  \resizebox{\linewidth}{!}{%
  \begin{tabular}{lccccccccccccccccccccc}
    \toprule
    Attack & \multicolumn{3}{c}{DwtDct} & \multicolumn{3}{c}{DwtDctSvd} & \multicolumn{3}{c}{RivaGAN} & \multicolumn{3}{c}{SSL Watermarking} & \multicolumn{3}{c}{StegaStamp} & \multicolumn{3}{c}{Stable Signature} & \multicolumn{3}{c}{Tree-Ring} \\
    \cmidrule(lr){2-4}\cmidrule(lr){5-7}\cmidrule(lr){8-10}\cmidrule(lr){11-13}\cmidrule(lr){14-16}\cmidrule(lr){17-19}\cmidrule(lr){20-22}
    & BER & PSNR & SSIM & BER & PSNR & SSIM & BER & PSNR & SSIM & BER & PSNR & SSIM & BER & PSNR & SSIM & BER & PSNR & SSIM & Acc Rem. & PSNR & SSIM \\
    \midrule
    Gaussian blur & 0.3481 & 29.30 & 0.8906 & 0.0000 & 30.36 & 0.9050 & 0.0000 & 30.77 & 0.8945 & 0.0000 & 28.16 & 0.8413 & 0.0000 & 28.09 & 0.8482 & 0.1813 & 30.53 & 0.9062 & 0.0000 & 29.48 & 0.9003 \\
    Gaussian noise & 0.6038 & 20.59 & 0.4996 & 0.3025 & 20.49 & 0.4762 & 0.0516 & 20.55 & 0.4794 & \textbf{0.5055} & 20.53 & 0.5135 & 0.0018 & 20.52 & 0.5072 & 0.1842 & 23.28 & 0.4925 & 0.0100 & 22.45 & 0.4555 \\
    JPEG & 0.4038 & \textbf{33.67} & \textbf{0.9673} & 0.0156 & \textbf{34.28} & \textbf{0.9680} & 0.0025 & \textbf{34.42} & \textbf{0.9653} & 0.0255 & \textbf{34.43} & \textbf{0.9515} & 0.0001 & \textbf{34.63} & \textbf{0.9544} & 0.0848 & \textbf{43.80} & \textbf{0.9950} & 0.0000 & \textbf{43.18} & \textbf{0.9923} \\
    Brightness & \textbf{0.4591} & 7.78 & 0.2871 & 0.3787 & 8.00 & 0.2931 & 0.1391 & 8.39 & 0.2910 & 0.0023 & 8.07 & 0.2538 & 0.0000 & 7.87 & 0.2584 & 0.3762 & 7.93 & 0.2425 & 0.0700 & 7.32 & 0.2448 \\
    Contrast & 0.4587 & 14.21 & 0.6466 & 0.3787 & 14.35 & 0.6687 & 0.1309 & 14.39 & 0.6497 & 0.0048 & 14.34 & 0.5788 & 0.0000 & 14.00 & 0.5811 & 0.4279 & 14.69 & 0.5733 & 0.0100 & 13.91 & 0.5498 \\
    Rotation & 0.4097 & 9.69 & 0.2187 & 0.4094 & 9.98 & 0.2392 & 0.3081 & 9.82 & 0.2329 & 0.0333 & 9.98 & 0.2122 & 0.4227 & 9.59 & 0.2100 & \textbf{0.4281} & 10.19 & 0.2255 & 0.7200 & 8.98 & 0.2101 \\
    Cropping & 0.0356 & 10.09 & 0.2795 & 0.0000 & 10.26 & 0.2909 & 0.0013 & 10.08 & 0.2674 & 0.4070 & 10.41 & 0.1865 & \textbf{0.4860} & 9.87 & 0.1823 & 0.1119 & 10.79 & 0.1939 & \textbf{1.0000} & 9.35 & 0.2446 \\
    \midrule
    Ours & 0.4106 & 31.69 & 0.9656 & \textbf{0.4856} & 31.54 & 0.9659 & \textbf{0.3697} & 31.65 & 0.9597 & 0.3207 & 29.43 & 0.9303 & 0.4251 & 31.37 & 0.9470 & 0.2275 & 28.35 & 0.8934 & 0.3200 & 26.81 & 0.8799 \\
    \bottomrule
  \end{tabular}}
\end{table*}

\begin{table*}[!b]
  \caption{Denoising, reconstruction, and regeneration attacks compared with our selected point. Bold BER is closest to $0.5$; bold detector removal and fidelity metrics are larger.}
  \label{tab:reconstruction-attack-comparison}
  \centering
  \scriptsize
  \resizebox{\linewidth}{!}{%
  \begin{tabular}{lccccccccccccccccccccc}
    \toprule
    Attack & \multicolumn{3}{c}{DwtDct} & \multicolumn{3}{c}{DwtDctSvd} & \multicolumn{3}{c}{RivaGAN} & \multicolumn{3}{c}{SSL Watermarking} & \multicolumn{3}{c}{StegaStamp} & \multicolumn{3}{c}{Stable Signature} & \multicolumn{3}{c}{Tree-Ring} \\
    \cmidrule(lr){2-4}\cmidrule(lr){5-7}\cmidrule(lr){8-10}\cmidrule(lr){11-13}\cmidrule(lr){14-16}\cmidrule(lr){17-19}\cmidrule(lr){20-22}
    & BER & PSNR & SSIM & BER & PSNR & SSIM & BER & PSNR & SSIM & BER & PSNR & SSIM & BER & PSNR & SSIM & BER & PSNR & SSIM & Acc Rem. & PSNR & SSIM \\
    \midrule
    BM3D & 0.4097 & \textbf{35.45} & 0.9422 & 0.3591 & \textbf{36.27} & 0.9412 & 0.0191 & \textbf{36.22} & 0.9391 & 0.2903 & \textbf{32.37} & 0.8715 & 0.0015 & \textbf{32.56} & 0.8837 & 0.2035 & \textbf{33.15} & \textbf{0.9019} & 0.0000 & \textbf{33.22} & \textbf{0.8973} \\
    CtrlRegen & \textbf{0.4216} & 21.16 & 0.6242 & 0.4713 & 21.58 & 0.6455 & \textbf{0.4803} & 21.66 & 0.6272 & 0.5298 & 20.83 & 0.5594 & \textbf{0.4505} & 20.91 & 0.5949 & 0.5112 & 21.57 & 0.6320 & 0.2200 & 21.68 & 0.6570 \\
    WatermarkAttacker & 0.4153 & 23.96 & 0.7154 & 0.4088 & 24.59 & 0.7357 & 0.4106 & 24.84 & 0.7210 & 0.4885 & 23.22 & 0.6439 & 0.2570 & 23.37 & 0.6755 & \textbf{0.5050} & 24.80 & 0.7378 & 0.1500 & 24.98 & 0.7702 \\
    SADRE-BMSHJ & 0.4113 & 29.30 & 0.8676 & 0.4000 & 29.96 & 0.8760 & 0.3934 & 30.15 & 0.8615 & \textbf{0.5053} & 28.31 & 0.8115 & 0.3683 & 28.41 & 0.8272 & 0.3360 & 30.34 & 0.8679 & 0.0200 & 29.91 & 0.8587 \\
    SADRE-Cheng & 0.4119 & 30.59 & 0.8943 & 0.3875 & 31.05 & 0.8968 & 0.4025 & 31.26 & 0.8851 & 0.4905 & 29.94 & 0.8503 & 0.3619 & 30.08 & 0.8629 & 0.2533 & 31.45 & 0.8861 & 0.0400 & 31.36 & 0.8812 \\
    SADRE-Diff60 & 0.4156 & 25.99 & 0.7663 & 0.4844 & 26.68 & 0.7844 & 0.3559 & 27.02 & 0.7647 & 0.4838 & 24.98 & 0.6906 & 0.2465 & 25.19 & 0.7237 & 0.4771 & 27.47 & 0.8087 & 0.0100 & 28.44 & 0.8559 \\
    \midrule
    Ours & 0.4106 & 31.69 & \textbf{0.9656} & \textbf{0.4856} & 31.54 & \textbf{0.9659} & 0.3697 & 31.65 & \textbf{0.9597} & 0.3207 & 29.43 & \textbf{0.9303} & 0.4251 & 31.37 & \textbf{0.9470} & 0.2275 & 28.35 & 0.8934 & \textbf{0.3200} & 26.81 & 0.8799 \\
    \bottomrule
  \end{tabular}}
\end{table*}

\clearpage
\section{Full Ablation Results}
\label{app:full-ablation}

The variants remove multiband magnitude alignment, the scheduled drop0 term, the pooled-VGG constraint, or the high-frequency residual input. The fg4 and fg16 variants change the Fourier-block group count from eight. The no-drop0 variant retains $\mathcal{L}_{\mathrm{recov}}$ on the same zero-auxiliary prediction and is therefore not a complete removal of zero-auxiliary supervision.

\begin{table*}[!b]
  \caption{Ablations on DwtDct, DwtDctSvd, RivaGAN, and SSL Watermarking at $k=1.10,u=0$. Bold BER is closest to $0.5$; bold PSNR/SSIM is larger.}
  \label{tab:abl-primary-methods}
  \centering
  \scriptsize
  \resizebox{\linewidth}{!}{%
  \begin{tabular}{lcccccccccccc}
    \toprule
    Variant & \multicolumn{3}{c}{DwtDct} & \multicolumn{3}{c}{DwtDctSvd} & \multicolumn{3}{c}{RivaGAN} & \multicolumn{3}{c}{SSL Watermarking} \\
    \cmidrule(lr){2-4}\cmidrule(lr){5-7}\cmidrule(lr){8-10}\cmidrule(lr){11-13}
    & BER & PSNR & SSIM & BER & PSNR & SSIM & BER & PSNR & SSIM & BER & PSNR & SSIM \\
    \midrule
    Full & 0.4106 & 31.69 & 0.9656 & 0.4856 & 31.54 & 0.9659 & 0.3697 & 31.65 & 0.9597 & 0.3208 & 29.43 & 0.9303 \\
    w/o spectral alignment & 0.4131 & \textbf{32.89} & \textbf{0.9772} & 0.4550 & \textbf{33.01} & \textbf{0.9776} & 0.2828 & \textbf{32.89} & \textbf{0.9713} & 0.2717 & \textbf{30.30} & \textbf{0.9517} \\
    w/o scheduled drop0 & \textbf{0.4225} & 27.95 & 0.9440 & \textbf{0.4928} & 27.33 & 0.9425 & \textbf{0.4934} & 27.52 & 0.9322 & \textbf{0.3610} & 26.76 & 0.8976 \\
    w/o perceptual constraint & 0.4144 & 31.35 & 0.9630 & 0.4684 & 31.18 & 0.9640 & 0.3159 & 31.11 & 0.9558 & 0.2995 & 30.00 & 0.9323 \\
    w/o high-frequency residual & 0.4147 & 31.69 & 0.9656 & 0.4891 & 31.58 & 0.9657 & 0.3669 & 31.54 & 0.9581 & 0.3090 & 29.57 & 0.9303 \\
    fg4 & 0.4163 & 31.51 & 0.9660 & 0.4794 & 31.29 & 0.9661 & 0.4006 & 31.37 & 0.9588 & 0.3093 & 29.99 & 0.9360 \\
    fg16 & 0.4125 & 31.62 & 0.9659 & 0.4822 & 31.48 & 0.9661 & 0.3678 & 31.59 & 0.9594 & 0.3313 & 29.44 & 0.9314 \\
    \bottomrule
  \end{tabular}}
\end{table*}

\begin{table*}[!b]
  \caption{Ablations on StegaStamp, Stable Signature, and Tree-Ring. Tree-Ring reports Acc Removal, not BER.}
  \label{tab:abl-learned-detector-methods}
  \centering
  \scriptsize
  \resizebox{\linewidth}{!}{%
  \begin{tabular}{lccccccccc}
    \toprule
    Variant & \multicolumn{3}{c}{StegaStamp} & \multicolumn{3}{c}{Stable Signature} & \multicolumn{3}{c}{Tree-Ring} \\
    \cmidrule(lr){2-4}\cmidrule(lr){5-7}\cmidrule(lr){8-10}
    & BER & PSNR & SSIM & BER & PSNR & SSIM & Acc Rem. & PSNR & SSIM \\
    \midrule
    Full & 0.4251 & 31.37 & 0.9470 & 0.2275 & 28.35 & 0.8934 & 0.3200 & 26.81 & 0.8799 \\
    w/o spectral alignment & 0.3573 & \textbf{33.28} & \textbf{0.9681} & 0.1975 & \textbf{28.78} & \textbf{0.9001} & 0.2700 & \textbf{27.54} & \textbf{0.8903} \\
    w/o scheduled drop0 & \textbf{0.4640} & 27.67 & 0.9175 & 0.2321 & 26.57 & 0.8825 & 0.1700 & 24.66 & 0.8410 \\
    w/o perceptual constraint & 0.4164 & 31.04 & 0.9415 & \textbf{0.3863} & 27.48 & 0.8360 & \textbf{0.3700} & 25.08 & 0.8579 \\
    w/o high-frequency residual & 0.4377 & 31.26 & 0.9453 & 0.2135 & 28.29 & 0.8922 & 0.3400 & 26.36 & 0.8756 \\
    fg4 & 0.4287 & 31.14 & 0.9471 & 0.2294 & 28.36 & 0.8955 & 0.3400 & 24.76 & 0.8702 \\
    fg16 & 0.4311 & 31.25 & 0.9461 & 0.2221 & 28.31 & 0.8950 & \textbf{0.3700} & 25.39 & 0.8758 \\
    \bottomrule
  \end{tabular}}
\end{table*}

\clearpage
\section{Full Double Residual Visualization}
\label{app:full-double-residual}

\begin{figure*}[!b]
  \centering
  \includegraphics[width=0.46\textwidth,height=0.70\textheight,keepaspectratio]{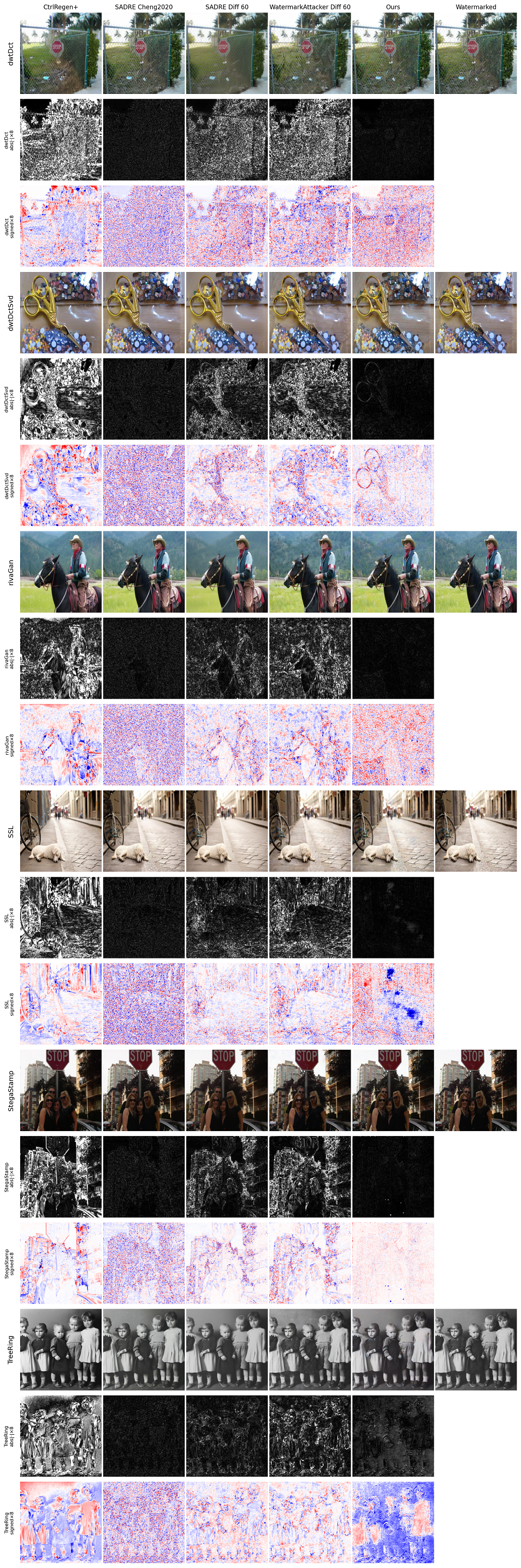}
  \caption{Full double residual comparison across all evaluated watermarking systems. Each watermarking block shows attack outputs, absolute residuals $|x_{\mathrm{atk}}-x_w|\times8$, and signed residuals $(x_{\mathrm{atk}}-x_w)\times8$ for the same attack methods as Tables~\ref{tab:traditional-attack-comparison} and~\ref{tab:reconstruction-attack-comparison}.}
  \label{fig:double-residual-full}
\end{figure*}

Figure~\ref{fig:double-residual-full} extends the diagnostic in Figure~\ref{fig:double-residual-sample} to all evaluated watermarking systems. The same absolute and signed residual scaling is used throughout. In these displayed samples, regeneration and diffusion methods produce broader bidirectional residuals. The proposed attack produces weaker residuals for several bit-decoder methods, while Tree-Ring shows broader signed changes. These images describe spatial modification patterns and do not identify watermark information or a causal frequency mechanism.

\clearpage
\section{Architecture Details}
\label{app:architecture}

The implementation uses an encoder $E$ and decoder $D$ with no normalization layers or dropout. Inputs are resized to $512\times512$ and normalized to $[-1,1]$. Every model output that is decoded as an image or latent image is also bounded by a final $\tanh$: $g\in[-1,1]^{1\times H\times W}$, $u\in[-1,1]^{C\times H\times W}$, and $D(g,u)\in[-1,1]^{3\times H\times W}$. The auxiliary channel count is fixed to $C=16$ in all reported runs.

Both branches use the same high frequency extractor,
\begin{equation}
  H(y)=y-G_{\sigma=1.0}(y),
  \label{eq:app-high-frequency-extractor}
\end{equation}
where $G$ is a $5\times5$ Gaussian blur with padding 2. For RGB inputs, $y=0.299R+0.587G+0.114B$; for one channel inputs, $y$ is used directly. The encoder input is $[x,H(\mathrm{rgb2gray}(x))]$ and therefore has four channels. The decoder input is $[g,u,H(g)]$ and therefore has $1+16+1=18$ channels.

\begin{table*}[!b]
  \caption{Encoder block table. A residual block is Conv$(3,1,1)$--ReLU--Conv$(3,1,1)$ plus an identity skip. A down block is Conv$(3,2,1)$--Conv$(3,1,1)$--ReLU. An up block is bilinear upsample by 2, Conv$(3,1,1)$--Conv$(3,1,1)$--ReLU, followed by skip addition.}
  \label{tab:app-encoder-blocks}
  \centering
  \small
  \begin{tabularx}{\textwidth}{p{0.20\textwidth}p{0.22\textwidth}p{0.18\textwidth}X}
    \toprule
    Stage & Operation & Output size & Notes \\
    \midrule
    Input fusion & Concatenate RGB and $H(\mathrm{rgb2gray}(x))$ & $4\times H\times W$ & High frequency extractor uses kernel 5 and $\sigma=1.0$. \\
    Head & Conv $4\!\rightarrow\!64$, ResBlock$(64)\times2$ & $64\times H\times W$ & First shared feature map. \\
    Down 1 & DownConv $64\!\rightarrow\!128$ & $128\times H/2\times W/2$ & The $64$ channel head output is stored as a skip. \\
    Down 2 & DownConv $128\!\rightarrow\!256$ & $256\times H/4\times W/4$ & The $128$ channel feature is stored as a skip. \\
    Bridge & ResBlock$(256)\times4$ & $256\times H/4\times W/4$ & Bottleneck refinement. \\
    Up 1 & UpConv $256\!\rightarrow\!128$ + skip & $128\times H/2\times W/2$ & Additive skip from Down 2 input. \\
    Up 2 & UpConv $128\!\rightarrow\!64$ + skip & $64\times H\times W$ & Additive skip from Down 1 input. \\
    Shared tail & ResBlock$(64)\times2$ & $64\times H\times W$ & Shared by the $g$ and $u$ heads. \\
    Fourier split & Fourier block & $64\times H\times W$ & Residual Fourier disentanglement block used in the reported model. \\
    \bottomrule
  \end{tabularx}
\end{table*}

The encoder has two output heads after the shared tail and optional Fourier split block:
\begin{align}
  g &= \tanh(\mathrm{Conv}_{32\rightarrow1}^{3\times3}(\mathrm{ReLU}(\mathrm{Conv}_{64\rightarrow32}^{3\times3}(f)))), \\
  u &= \tanh(\mathrm{Conv}_{64\rightarrow16}^{3\times3}(\mathrm{ReLU}(\mathrm{Conv}_{64\rightarrow64}^{3\times3}(f)))) .
\end{align}
Thus the structural head is $64\rightarrow32\rightarrow1$ and the auxiliary head is $64\rightarrow64\rightarrow16$. Both heads use stride 1 and padding 1 for all $3\times3$ convolutions.

\begin{table*}[!b]
  \caption{Decoder block table. The decoder is full resolution throughout; it does not downsample.}
  \label{tab:app-decoder-blocks}
  \centering
  \small
  \begin{tabularx}{\textwidth}{p{0.22\textwidth}p{0.24\textwidth}p{0.18\textwidth}X}
    \toprule
    Stage & Operation & Output size & Notes \\
    \midrule
    Input fusion & Concatenate $g$, $u$, and $H(g)$ & $18\times H\times W$ & $C=16$, so $1+C+1=18$. \\
    Projection & Conv $18\!\rightarrow\!64$ & $64\times H\times W$ & $3\times3$, stride 1, padding 1. \\
    Residual trunk & ResBlock$(64)\times8$ & $64\times H\times W$ & Full resolution residual refinement. \\
    Expansion & Conv $64\!\rightarrow\!256$, GELU & $256\times H\times W$ & $3\times3$, stride 1, padding 1. \\
    RGB head & Conv $256\!\rightarrow\!3$, $\tanh$ & $3\times H\times W$ & $1\times1$ projection to RGB in $[-1,1]$. \\
    \bottomrule
  \end{tabularx}
\end{table*}

\paragraph{Frequency blocks.}
The encoder frequency block is a residual rFFT block operating on the 64 channel shared feature map. It applies a two-dimensional orthonormal real Fourier transform, splits the real and imaginary components into 8 token groups, and therefore uses 8 channels per group. Each group has a learned real mixing matrix and imaginary mixing matrix of shape $8\times8$, initialized with standard deviation 0.02. Learned per-channel real and imaginary scales of shape $1\times64\times1\times1$ are applied, followed by a soft-thresholding nonlinearity with threshold 0.01, inverse rFFT, and residual addition to the input feature.

The latent attack operator $A_g$ is also frequency-domain by default. It applies rFFT to $g_w$, builds normalized radial masks with low band $r\le 1/3$, mid band $1/3<r\le2/3$, and high band $r>2/3$, attenuates each band, adds magnitude noise on the mid/high bands, adds phase noise on the high band, reconstructs with inverse rFFT, and clamps the result to $[-1,1]$. Stage 2 keep ratios are $(0.95,0.75,0.60)$ for low/mid/high, with magnitude and phase noise $\sigma=(0.02,0.03)$. Stage 3 keep ratios are $(0.90,0.55,0.35)$, with magnitude and phase noise $\sigma=(0.05,0.06)$. The ring-notch option is disabled for the main $(k,u)$ results and module ablations; it is enabled only for the Tree-Ring detector-specific radius scan in Table~\ref{tab:treering-radius-scan}.

\clearpage
\section{Full Loss Weights}
\label{app:full-loss-weights}

The training objective is
\begin{equation}
\begin{aligned}
  \mathcal{L} ={}&
  \lambda_{\mathrm{inv}}\mathcal{L}_{\mathrm{inv}}
  + \lambda_{\mathrm{grayc}}\mathcal{L}_{\mathrm{grayc}}
  + \lambda_{\mathrm{rec1}}\mathcal{L}_{\mathrm{rec1}} \\
  &+ \lambda_{\mathrm{g-atk}}\mathcal{L}_{\mathrm{g-atk}}
  + \lambda_{\mathrm{gray\_hatw}}\mathcal{L}_{\mathrm{gray\_hatw}} \\
  &+ w_{\mathrm{drop0}}(e)\mathcal{L}_{\mathrm{drop0}}
  + w_{\mathrm{pvc0}}(e)\mathcal{L}_{\mathrm{pvc0}}
  + \lambda_{\mathrm{recov}}\mathcal{L}_{\mathrm{recov}} \\
  &+ w_{\mathrm{pvc\_pred0}}(e)\mathcal{L}_{\mathrm{pvc\_pred0}} \\
  &+ \lambda_{\mathrm{hf\_pred0}}\mathcal{L}_{\mathrm{hf\_pred0}}
  + \lambda_{\mathrm{edge\_pred0}}\mathcal{L}_{\mathrm{edge\_pred0}} \\
  &+ \lambda_{\mathrm{uc}}\|u_c\|_1 \\
  &+ \lambda_{\mathrm{quant\_hatw}}\mathcal{L}_{\mathrm{quant\_hatw}} \\
  &+ \lambda_{\mathrm{g\_hf\_hatw}}\mathcal{L}_{\mathrm{g\_hf\_hatw}}
  + \lambda_{\mathrm{g\_edge\_hatw}}\mathcal{L}_{\mathrm{g\_edge\_hatw}} \\
  &+ \lambda_{\mathrm{fft\_split}}\mathcal{L}_{\mathrm{fft\_split}} \\
  &+ \lambda_{\mathrm{uw\_floor}}\mathcal{L}_{u\mathrm{floor}}
  + \lambda_{\mathrm{clean0}}\mathcal{L}_{\mathrm{clean0}} .
\end{aligned}
\label{eq:appendix-code-objective}
\end{equation}
Table~\ref{tab:app-loss-weights} lists every fixed or scheduled loss weight used by the training schedule in Appendix~\ref{app:training-schedule}. The implementation retains two pairs of duplicate Stage~3 computations for historical scheduling reasons. Specifically, $\mathcal{L}_{\mathrm{drop0}}$ and $\mathcal{L}_{\mathrm{recov}}$ are both $\|x_{\mathrm{pred0}}-x_c\|_1$; after warmup their coefficients add to an effective pixel weight of $10+6=16$. Likewise, $\mathcal{L}_{\mathrm{pvc0}}$ and $\mathcal{L}_{\mathrm{pvc\_pred0}}$ reuse the same pooled-VGG value and have an effective Stage~3 weight of $4+6=10$ after perceptual warmup. Reporting these effective weights makes the objective and existing checkpoints easier to interpret; full reproducibility still requires the exact run configuration, random seeds, data manifest, evaluation commands, and per-variant checkpoints listed in Section~\ref{sec:ablation}. Future implementations should consolidate each pair before running causal ablations.

\begin{table*}[!b]
  \caption{Complete loss weights used in the reported training schedule. Stage 1, Stage 2, and Stage 3 correspond to epochs 0--14, 15--79, and 80--139.}
  \label{tab:app-loss-weights}
  \centering
  \footnotesize
  \renewcommand{\arraystretch}{0.92}
  \begin{tabularx}{\textwidth}{>{\raggedright\arraybackslash}p{0.16\textwidth}>{\raggedright\arraybackslash}p{0.12\textwidth}>{\raggedright\arraybackslash}p{0.12\textwidth}>{\raggedright\arraybackslash}X}
    \toprule
    Weight & Value & Active stage & Loss term / note \\
    \midrule
    $\lambda_{\mathrm{inv}}$ & 2.0 & 1--3 & MSE between $x_c$ and $D(g_c,u_c)$. \\
    $\lambda_{\mathrm{grayc}}$ & 1.0 & 1--3 & Grayscale conformity for $g_c$. \\
    $\lambda_{\mathrm{rec1}}$ & 1.0 & 1--3 & Multiplies $\|D(g_w,u_w)-x_w\|_1$ after the schedule scale $s_{\mathrm{rec1}}(e)$. \\
    $\lambda_{\mathrm{g-atk}}$ & 3.0 & 2--3 & $\|A_g(g_w;k)-g_c\|_1$; zero in Stage 1. \\
    $\lambda_{\mathrm{gray\_hatw}}$ & 1.0 & 1--3 & Grayscale conformity for the attacked structural latent. In Stage 1, $A_g$ is identity. \\
    $\lambda_{\mathrm{drop0}}$ & target 10.0 & 2--3 & Same Stage~3 residual as $\mathcal{L}_{\mathrm{recov}}$; warms from 2.0 to 10.0 in Stage~2 and is 10.0 in Stage~3. \\
    $\lambda_{\mathrm{pvc0,s1}}$ & 0.0 & 1 & Pooled VGG cosine weight for $D(A_g(g_w),0)$ in Stage 1. \\
    $\lambda_{\mathrm{pvc0,s2}}$ & 2.0 & 2 & Multiplied by $\min(1,3p_{\mathrm{s2}})$, where $p_{\mathrm{s2}}$ is Stage 2 progress. \\
    $\lambda_{\mathrm{pvc0,s3}}$ & 4.0 & 3 & Reuses the same Stage~3 pooled-VGG value as $\mathcal{L}_{\mathrm{pvc\_pred0}}$ and follows the five-epoch warmup. \\
    $\lambda_{\mathrm{recov}}$ & 6.0 & 3 & Same $\|x_{\mathrm{pred0}}-x_c\|_1$ residual as $\mathcal{L}_{\mathrm{drop0}}$. \\
    $\lambda_{\mathrm{pvc\_pred0}}$ & 6.0 & 3 & Reuses $\mathcal{L}_{\mathrm{pvc0}}$ in Stage~3 and follows the same five-epoch warmup. \\
    $\lambda_{\mathrm{hf\_pred0}}$ & 0.8 & 3 & Multiscale high frequency loss on $x_{\mathrm{pred0}}$. \\
    $\lambda_{\mathrm{edge\_pred0}}$ & 0.5 & 3 & Sobel edge consistency on $x_{\mathrm{pred0}}$. \\
    $\lambda_{\mathrm{uc}}$ & 1.0 & 1--3 & $\|u_c\|_1$ clean auxiliary sparsity. \\
    $\lambda_{\mathrm{quant\_hatw}}$ & 1.0 & 3 & Quantization loss on $A_g(g_w;k)$ toward the nearest 8-bit level in $[-1,1]$. \\
    $\lambda_{\mathrm{g\_hf\_hatw}}$ & 0.3 & 3 & Multiscale high frequency branch loss for $(g_c,A_g(g_w;k))$. \\
    $\lambda_{\mathrm{g\_edge\_hatw}}$ & 0.2 & 3 & Sobel branch loss for $(g_c,A_g(g_w;k))$. \\
    $\lambda_{\mathrm{fft\_split}}$ & 1.2 & 2--3 & Multiband Fourier magnitude alignment between $A_g(g_w;k)$ and $g_c$. \\
    $\lambda_{\mathrm{uw\_floor}}$ & 0.2 & 2--3 & $\max(0,\tau-\mathbb{E}|u_w|)$ with $\tau=0.04$. \\
    $\lambda_{\mathrm{clean0}}$ & 0.0 & disabled & Clean zero-auxiliary anchor branch is not computed. \\
    $\lambda_{\mathrm{gc\_hf}}$ & 1.0 & inner weight & Internal clean structural high frequency / edge branch weight. \\
    $\lambda_{\mathrm{gw\_hf}}$ & 0.5 & inner weight & Internal attacked structural high frequency / edge branch weight. \\
    \bottomrule
  \end{tabularx}
\end{table*}

The remaining scalar loss constants are as follows. Grayscale conformity uses threshold $70/127$, VGG-19 features up to layer 21, and contrast weight $10^{-7}$. Its local structure weight is 0.5 in Stage 1, 0.2 in Stage 2, and 0.1 in Stage 3. The multiscale high frequency kernels are $(3,5,9)$ with sigmas $(0.8,1.2,2.0)$ and scale weights $(0.5,1.0,1.5)$. The Fourier separation loss uses low/mid/high band weights $(0.8,1.0,1.0)$. Charbonnier losses use $\epsilon=10^{-3}$.

\clearpage
\section{Training Schedule}
\label{app:training-schedule}

Table~\ref{tab:app-stage-transition} records the epoch ranges and the quantities that change at each transition.

\begin{table*}[!b]
  \caption{Stage transition and scheduled quantities.}
  \label{tab:app-stage-transition}
  \centering
  \footnotesize
  \begin{tabularx}{\textwidth}{>{\raggedright\arraybackslash}p{0.12\textwidth}>{\raggedright\arraybackslash}p{0.12\textwidth}>{\raggedright\arraybackslash}X}
    \toprule
    Stage & Epochs & Schedule \\
    \midrule
    Stage 1 & 0--14 & $A_g$ is identity, attack strength $k=0$, zero auxiliary attack losses are inactive, and $s_{\mathrm{rec1}}=1$. \\
    Stage 2.1 & 15--35 & Stage 2 starts. Attack strength follows a cosine ramp from 0.25 toward 0.75, $\lambda_{\mathrm{drop0}}$ warms from 2.0 toward 10.0, and $s_{\mathrm{rec1}}$ begins cosine decay. \\
    Stage 2.2 & 36--57 & Middle third of Stage 2; same schedules continue. The pooled VGG Stage 2 scale is $\min(1,3p_{\mathrm{s2}})$ and is saturated after the first third of Stage 2. \\
    Stage 2.3 & 58--79 & Final third of Stage 2. Attack strength reaches 0.75 and $s_{\mathrm{rec1}}$ reaches its floor 0.2. \\
    Stage 3 & 80--139 & Full objective is active. Attack strength ramps continuously from 0.75 to 1.0 during the first 8 Stage 3 epochs and stays at 1.0 afterwards. Perceptual Stage 3 weights warm up over the first 5 Stage 3 epochs. \\
    \bottomrule
  \end{tabularx}
\end{table*}

\clearpage
\section{Additional evaluation protocol}
\label{app:evaluation}

\subsection{Parameter search}

The coarse grid evaluates $k\in\{0,0.25,0.50,0.75,1.00\}$ and $u\in\{0,0.25,0.50,0.75,1.00\}$. The fine grid refines the high performing region with $k\in[0.80,1.20]$ and $u\in\{0,0.03,0.05,0.06,0.10,0.15,0.20\}$. The shared selected point is chosen by maximizing the attack metric subject to visual quality constraints, rather than by maximizing BER alone. This distinction matters because destructive attacks such as brightness shifts, rotation, and cropping can produce high detector failure while making the output visibly unusable.

\subsection{Metric interpretation}

For bit-decoder watermarks, BER is computed between the original embedded payload and the payload extracted after attack. Stronger removal means BER closer to $0.5$, summarized by $\mathrm{RR}=1-2|\mathrm{BER}-0.5|$; BER above $0.5$ is not automatically stronger. For Tree-Ring, lower TPR at the nominal 1\%-FPR calibration target means stronger removal, and Acc Removal is $1-\mathrm{TPR}$. The realized empirical FPR should be reported when it differs from the nominal target. PSNR and SSIM are computed against the watermarked input unless otherwise specified, so they measure visible preservation under attack rather than recovery of an unavailable clean image.

\end{document}